\documentclass[10pt,twocolumn,letterpaper]{article}

\usepackage{iccv}        

\definecolor{iccvblue}{rgb}{0.21,0.49,0.74}
\usepackage[pagebackref,breaklinks,colorlinks,allcolors=iccvblue]{hyperref}
\usepackage{algorithm}
\usepackage{algorithmic}
\usepackage{graphicx}
\usepackage{array}
\usepackage{booktabs}
\usepackage{multirow}
\usepackage{adjustbox}
\usepackage[singlelinecheck=false]{caption}
\usepackage{pgfplots}
\usepackage{subcaption}
\usepackage{amsmath}
\usepackage{amssymb}
\usepackage{mathtools}
\usepackage{amsthm}
\usepackage{xspace}
\usepackage{overpic}
\theoremstyle{plain}

\theoremstyle{definition}
\newtheorem{definition}{Definition}[section]

\theoremstyle{remark}

\usepackage[font=small,labelfont=bf,justification=centering]{caption}
\usepackage{subcaption}
\usepackage{xcolor}
\definecolor{DeepSkyBlue}{RGB}{0,191,255}
\usepackage{placeins} 
\usepackage{stfloats} 
\usepackage{float}   
\usepackage{lscape}  
\usetikzlibrary{intersections}
\usepgfplotslibrary{fillbetween}
\usepackage{newtxtext,newtxmath}
\usetikzlibrary{calc}
\def\paperID{4056} 
\def\confName{ICCV}
\def\confYear{2025}

\title{Accelerating Diffusion Sampling via Exploiting Local Transition Coherence}

\author{
    Shangwen Zhu\thanks{These authors contributed equally to this work.}, 
    Han Zhang\footnotemark[1], 
    Zhantao Yang\footnotemark[1], Qianyu Peng\footnotemark[1], Zhao Pu\\
    Huangji Wang, 
    Fan Cheng\thanks{Corresponding author}\\
    Shanghai Jiao Tong University, Shanghai, China\\
    \footnotesize
    \texttt{{zhushangwen6,hzhang9617,ztyang196,qianyupeng312}@gmail.com}
    \\
    \footnotesize\texttt{{wanghuangji,pz-23dy}@sjtu.edu.cn}
    \quad \footnotesize\texttt{chengfan85@gmail.com}
}

\newcommand{\methodabbr}{\textsc{LTC-Accel}\xspace}
\newcommand{\phenmn}{\text{Local Transition Coherence}\xspace}

\definecolor{myblue}{RGB}{0,114,189}     
\definecolor{myorange}{RGB}{217,83,25}   
\definecolor{mygreen}{RGB}{119,172,48}   
\definecolor{myred}{RGB}{162,20,47}      
\definecolor{mypurple}{RGB}{126,47,142}  

\definecolor{Red0}{HTML}{FFF5F5}  
\definecolor{Red1}{HTML}{FFDADA}  
\definecolor{Red2}{HTML}{FFA3A3}  
\definecolor{Red3}{HTML}{FF6B6B}  
\definecolor{Red4}{HTML}{E03131}  
\definecolor{Red5}{HTML}{A61D1D}  
\definecolor{Red6}{HTML}{330000}  

\definecolor{Blue1}{RGB}{135,206,235}
\definecolor{Blue2}{RGB}{135,206,235}
\definecolor{Blue3}{RGB}{42,82,190}
\definecolor{Blue4}{RGB}{0,71,171}
\definecolor{Blue5}{RGB}{0,0,128}
\definecolor{Blue6}{RGB}{25,25,112}

\definecolor{crystal-blue}{RGB}{65, 130, 210}
\definecolor{lavender-dream}{RGB}{140, 95, 200}
\definecolor{velvet-ruby}{RGB}{145, 25, 45}
\definecolor{ember-glow}{RGB}{210, 60, 80}  

\definecolor{plot-blue}{RGB}{0, 105, 120}

\begin{document}
\maketitle


\newcommand{\tocite}[1]{\textcolor{red}{[TO CITE]}}


\begin{abstract}
Text-based diffusion models have made significant breakthroughs in generating high-quality images and videos from textual descriptions. However, the lengthy sampling time of the denoising process remains a significant bottleneck in practical applications. Previous methods either ignore the statistical relationships between adjacent steps or rely on attention or feature similarity between them, which often only works with specific network structures. To address this issue, we discover a new statistical relationship in the transition operator between adjacent steps, focusing on the relationship of the outputs from the network. This relationship does not impose any requirements on the network structure. Based on this observation, we propose a novel \textbf{training-free} acceleration method called \methodabbr, which uses the identified relationship to estimate the current transition operator based on adjacent steps. Due to no specific assumptions regarding the network structure, \methodabbr is applicable to almost all diffusion-based methods and orthogonal to almost all existing acceleration techniques, making it easy to combine with them. Experimental results demonstrate that \methodabbr significantly speeds up sampling in text-to-image and text-to-video synthesis while maintaining competitive sample quality. Specifically, \methodabbr achieves a speedup of $\mathbf{1.67\times}$ in Stable Diffusion v2 and a speedup of $\mathbf{1.55\times}$ in video generation models. When combined with distillation models, \methodabbr achieves a remarkable $\mathbf{10\times}$ speedup in video generation, allowing \textbf{real-time} generation of more than $\mathbf{16 \text{FPS}}$. Our code (include colab version) is available on \href{https://zhushangwen.github.io/LTC-accel.io/}{Project Page}.
\end{abstract}

\section{Introduction}
\label{sec:intro}

\begin{figure}[ht]
        \centering
        \includegraphics[width=0.45\textwidth]{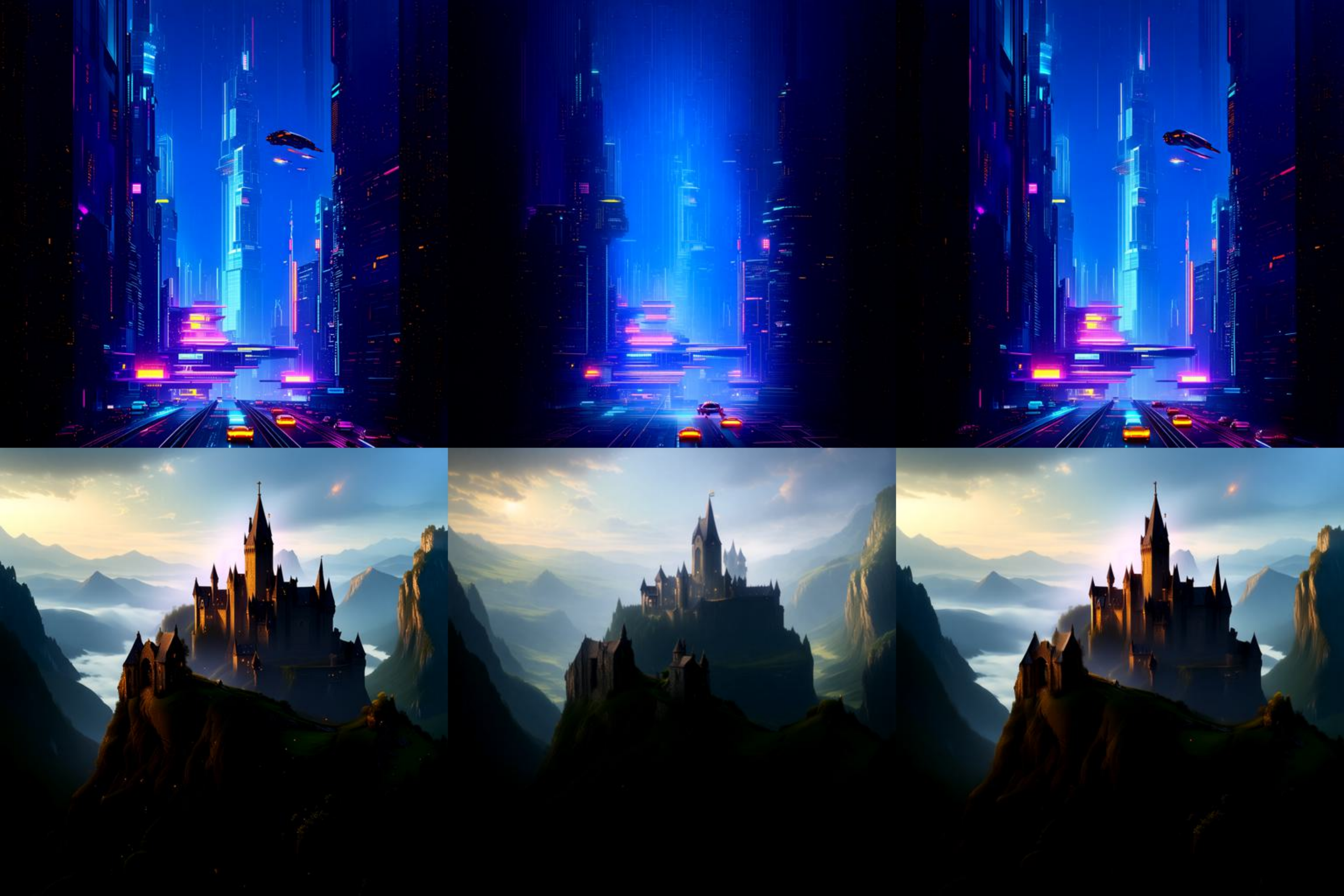}
        \begin{minipage}{\linewidth}
            \centering
            \hspace*{0.5pt}
            \begin{minipage}{0.3\linewidth}
            \centering
            \footnotesize Baseline\\$12$ steps
            \end{minipage}
            \hspace*{2pt}
            \begin{minipage}{0.3\linewidth}
            \centering
            \footnotesize  DPM-Solver++\\$8$ steps
            \end{minipage}
            \hspace*{4pt}
            \begin{minipage}{0.32\linewidth}
            \centering
            \footnotesize\methodabbr (Ours)\\$8$ steps ($\mathbf{1.5\times}$speedup)
            \end{minipage}
            \hspace*{\fill}
        \end{minipage}
        \begin{minipage}{0.35\linewidth}
            \vspace*{-2.4\linewidth}
            \hspace{-\linewidth}
            \rotatebox[origin=c]{90}{ 
                \parbox{1.2\linewidth}{
                    \centering 
                    Stable Diffusion v3.5
                }
            }
        \end{minipage}
        \vspace{-5pt}
        \captionsetup{justification=justified, singlelinecheck=false}
        \caption{Comparison between \methodabbr with DPM-Solver++ when implemented on Stable Diffusion v3.5 under the same number of steps and speedup framework. Results show that \methodabbr significantly outperforms DPM-Solver++.}
        \vspace{-12pt}
        \label{fig:1_Comparision on sd35}
\end{figure}

\begin{figure*}[t]
    \centering
    \begin{subfigure}{0.45\textwidth}
        \centering
        \includegraphics[width=\textwidth]{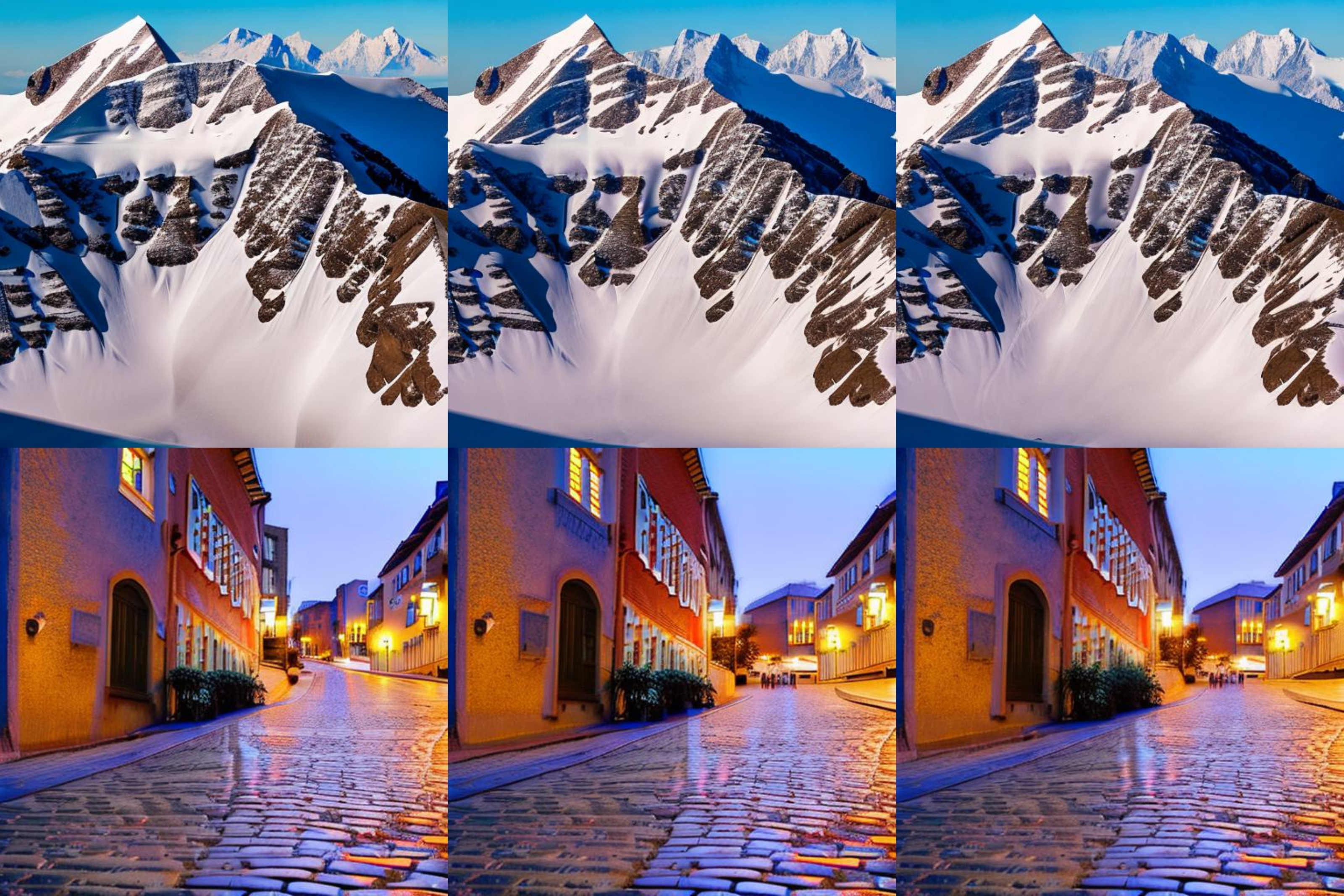}
        \begin{minipage}{\linewidth}
            \centering
            \hspace*{\fill}
            \begin{minipage}{0.3\linewidth}
            \centering
            \footnotesize Baseline\\$50$ steps
            \end{minipage}
            \hspace*{\fill}
            \begin{minipage}{0.3\linewidth}
            \centering
            \footnotesize Cache-based method\\$50$ steps\\($1.66\times$ speedup)
            \end{minipage}
            \hspace*{\fill}
            \begin{minipage}{0.3\linewidth}
            \centering
            
           \footnotesize \methodabbr+Cache\\$38 $steps\\($\mathbf{2.34\times}$ speedup)
            \end{minipage}
            \hspace*{\fill}
        \end{minipage}
        \begin{minipage}{0.37\linewidth}
            \vspace*{-2.5\linewidth}
            \hspace{-\linewidth}
            \rotatebox[origin=c]{90}{ 
                \parbox{\linewidth}{
                    \centering 
                    Stable Diffusion v2
                }
            }
        \end{minipage}
        \caption{Combing \methodabbr with Deepcache (Cache).}
        \label{fig:1_cache on sd v2}
    \end{subfigure}
    \hspace{0.02\textwidth}
    \begin{subfigure}{0.45\textwidth}
        \centering
        \includegraphics[width=\textwidth]{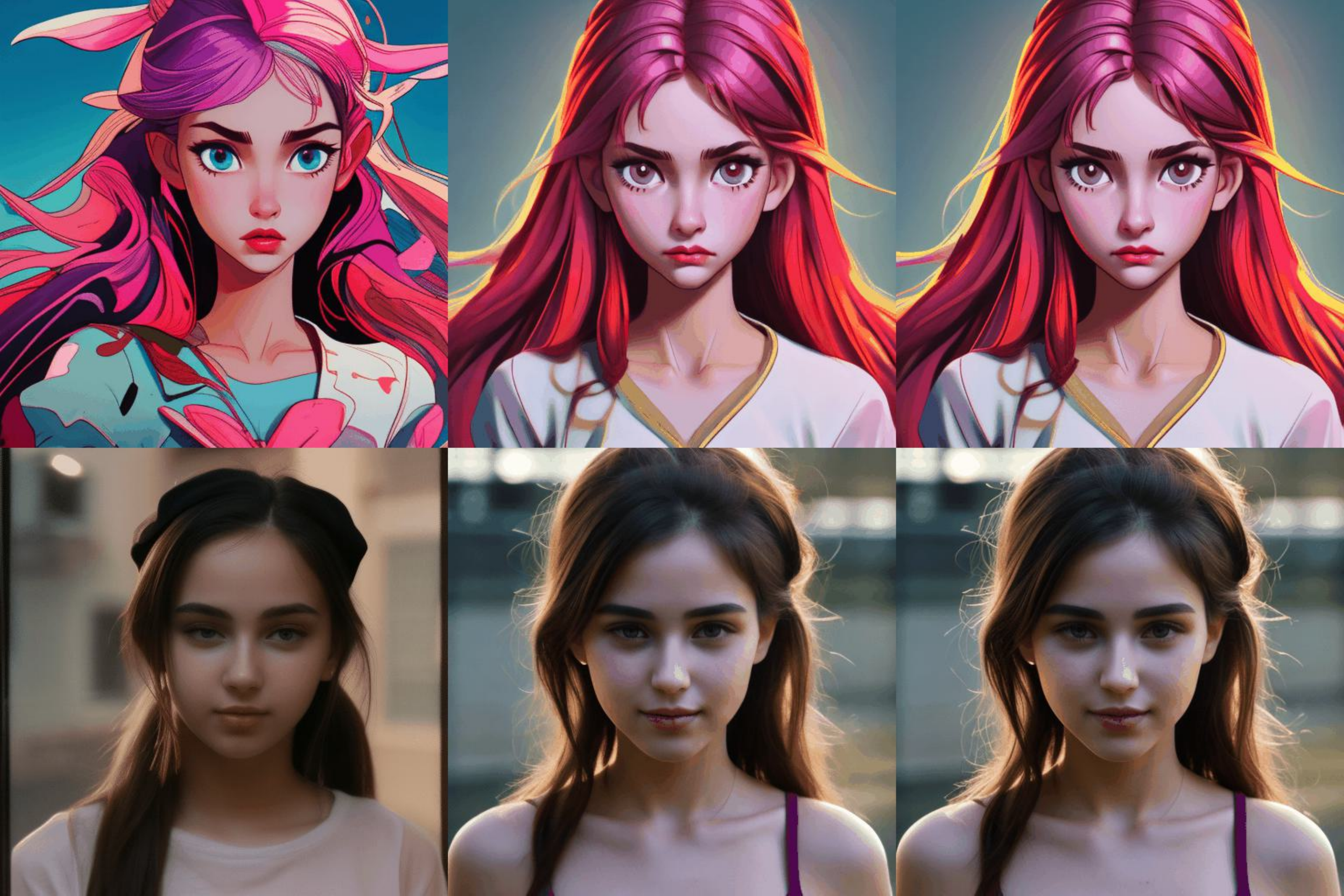}
        \begin{minipage}{\linewidth}
            \centering
            \hspace*{\fill}
            \begin{minipage}{0.3\linewidth}
            \centering
            \footnotesize Baseline\\$30$ steps
            \end{minipage}
            \hspace*{\fill}
            \begin{minipage}{0.3\linewidth}
            \centering
             \footnotesize Distillation \\$4 $steps\\($7.5\times$ speedup)
            \end{minipage}
            \hspace*{\fill}
            \begin{minipage}{0.3\linewidth}
            \centering
             \footnotesize
            \methodabbr + Distill \\$3$ steps\\($\mathbf{10\times}$ speedup)
            \end{minipage}
            \hspace*{\fill}
        \end{minipage}
        \begin{minipage}{0.37\linewidth}
            \vspace*{-2.5\linewidth}
            \hspace{-\linewidth}
            \rotatebox[origin=c]{90}{ 
                \parbox{1.1\linewidth}{
                    \centering 
                    Animated-Diff
                }
            }
        \end{minipage}
        \caption{Combing with Distillation Model (Distill).}
        \label{fig:1_Comparison with training-free approaches on anime-diff}
    \end{subfigure}
    \captionsetup{justification=justified, singlelinecheck=false}
    \caption{Qualitative results of \methodabbr integrated with existing training-free and training-based methods. \cref{fig:1_cache on sd v2}: Integration of \methodabbr with caching-based methods using DDIM on Stable Diffusion v2. \cref{fig:1_Comparison with training-free approaches on anime-diff}: Integration of \methodabbr with Animated-Diff-Lightning (distilled version of Animated-Diff). Results show that \methodabbr can be combined with previous methods well and achieve additional speedup without compromise on the quality of the generated images.
    }
    \label{}
\end{figure*}
Recent advancements in text-based generation, particularly with diffusion models~\citep{ho2020denoising, pmlr-v139-nichol21a, podell2023sdxl, karras2024analyzing, peebles2023scalable}, have significantly improved the generation of high-fidelity images~\citep{dhariwal2021diffusion, 0Rich, 2023MobileDiffusion, 2024Chest, S2025Grounding, 2024From}, audio~\citep{kong2021diffwaveversatilediffusionmodel, lu2022conditional, li2024diffusion, richter2023speech, lemercier2023storm, trachu2024thunder, welker2022speech}, and video~\citep{yang2024cogvideox, guo2023animatediff, lin2024animatediff, 2024Grid, 2025On, 2023Diffusion, 2025SignGen, 2025MEVG} from textual descriptions, achieving remarkable advancements in visual fidelity and semantic alignment. By iteratively refining a noisy input until it converges to a sample that aligns with the given text prompt, these models capture intricate details and complex compositions previously unattainable with other approaches. Despite their impressive capabilities, a major drawback of diffusion models, particularly in video generation, is their high computational complexity during the denoising process, leading to prolonged inference times and substantial computational costs. For instance, generating a $5$-second video at $8$ frames per second (FPS) with a resolution of $720\text{P}$ using Wan2.1-14B~\citep{wan2.1} on a single H$20$ GPU takes approximately $\mathbf{6935}$ seconds, highlighting the significant resource demands of high-quality video synthesis. This limitation poses a considerable challenge for real-time applications and resource-constrained environments~\citep{song2020denoising, dhariwal2021diffusion}.

To accelerate diffusion models, various approaches have been proposed, including training-based and training-free strategies. Training-based methods enhance sampling efficiency by modifying the training process~\citep{zhou2024score,yin2024one,sauer2024adversarial,salimans2022progressive,kim2023consistency,luo2023latent,song2023consistency} or altering model architectures~\citep{2018MobileNetV2,zhang2018shufflenet,krishnamoorthi2018quantizing,2020Designing}, but they require additional computation and extended training. In contrast, training-free methods improve sampling efficiency without modifying the trained model by optimizing the denoising process~\citep{karras2025guiding,qian2024boosting,xue2024accelerating,ho2022classifier,sabour2024align} or introducing more efficient solvers~\citep{zheng2023dpm,lu2022dpm,lu2022dpm+}. Additionally, caching-based methods such as DeepCache~\citep{2023DeepCache, wimbauer2024cache} exploit temporal redundancy in denoising steps to store and reuse intermediate features, thus achieving the reduction of redundant computations. However, these methods require a redesign of the caching strategy when the network architecture changes, limiting their flexibility across different models and configurations. Furthermore, they usually necessitate extra memory overhead to store intermediate representations, imposing constraints on resource-constrained deployment scenarios.

Unlike previous methods that rely on attention mechanisms or feature similarity within the network, we identify the phenomenon of \textbf{\phenmn}, which refers to the strong correlation between the transition operators ($\Delta x_{t+1, t}$) of neighboring steps. Based on this insight, we propose \methodabbr, a novel training-free acceleration method that approximates the current step's transition operator using those from adjacent steps. As a result, it does not depend on any specific network architecture, making it broadly applicable to various diffusion models and compatible with both training-based and training-free acceleration methods.

We conducted extensive experiments demonstrating the effectiveness of our method and its compatibility with other approaches. \methodabbr achieves a $\textbf{1.67}\times$ speedup on Stable Diffusion v2~\citep{Rombach_2022_CVPR}, and when combined with DeepCache~\citep{2023DeepCache}, accelerates the process to $\mathbf{2.34\times}$. It also integrates with Align Your Steps~\citep{sabour2024align}, achieving the equivalent of $10$ steps of Align Your Steps in just $8$ steps on Stable Diffusion v1.5~\citep{Rombach_2022_CVPR}, with minimal impact on generation quality. Additionally, we achieve a $\mathbf{1.67\times}$ speedup on the Animated-Diff model~\citep{guo2023animatediff}, and by combining it with the distilled version, Animated-Diff-Lightning~\citep{lin2024animatediff}, we achieve a $\mathbf{10\times}$ speedup in video generation, enabling \textbf{three-step} generation. These results collectively illustrate that our proposed approach not only significantly reduces the computational load but also synergizes effectively with other optimization methods, facilitating highly efficient inference even under stringent resource constraints.  

The core contributions of our work are:
\begin{itemize}
    \item  We have identified the phenomenon of Local Transition Coherence. Unlike previous approaches that rely on attention mechanisms or feature similarity within the network, this phenomenon reveals the inherent consistency of update trajectories, which is a broader consistency independent of any specific network architecture and pervasive throughout the diffusion sampling process.
    \item  We propose \methodabbr, a training-free, highly generalizable method applicable to various diffusion models and orthogonal to both training-based and training-free acceleration methods, providing significant acceleration without sacrificing performance.
    \item We conducted extensive experiments to validate the effectiveness of our method and its compatibility with other approaches. \methodabbr significantly accelerates the generation process of models including Stable Diffusion, CogVideoX and Animated-Diff, and enhances performance when combined with existing acceleration methods.
\end{itemize}
\section{Related Work}
\label{sec:relatedwork}

\begin{figure*}[t]
    \centering
    \hspace{-0.06\textwidth}
    \begin{subfigure}{0.28\textwidth}
        \centering
        \begin{tikzpicture}[scale=0.7]
        \begin{axis}[
            xlabel={Step},
            ylabel={Angle},
            xlabel style={xshift=20pt, font=\fontsize{10}{12}\selectfont\bfseries},
            ylabel style={yshift=-10pt, xshift=15pt, font=\fontsize{10}{12}\selectfont\bfseries},
            xmin=0, xmax=40,
            ymin=0, ymax=0.5,
            ytick={0.1,0.2,0.3,0.4,0.5},
            grid=major,
            grid style={line width=0.3pt, draw=gray!30},
            legend style={
                at={(0.95,1.4)}, 
                anchor=north, 
                font=\fontsize{9}{11}\selectfont, 
                draw=none, 
                fill=white, 
                fill opacity=0.9,
                legend cell align=left,
                legend image code/.code={   
                    \draw[draw={rgb,255:red,15;green,82;blue,186}, line width=1pt]
                    (0cm,0cm) -- (0.6cm,0cm);  
                },
            },
            every axis plot/.append style={thick},
            tick label style={font=\fontsize{9}{11}\selectfont}
        ]
    
        \addplot[name path=max_curve, draw=none] table[x=Timestep, y=Angle, col sep=comma] {sec/Data/angle_max.csv};
        \addplot[name path=min_curve, draw=none] table[x=Timestep, y=Angle, col sep=comma] {sec/Data/angle_min.csv};
    
        \addplot[color={rgb,255:red,15;green,82;blue,186}, fill opacity=0.2] fill between[of=max_curve and min_curve];
    
        \addplot[color={rgb,255:red,15;green,82;blue,186}, thick, smooth] table[x=Timestep, y=Angle, col sep=comma] {sec/Data/angle_mean.csv};
        \addlegendentry{Mean Angle}
    
        \end{axis}
        \end{tikzpicture}
        \vspace*{-8pt}
        \put(-30, 15){(\textbf{a}) Angle variation example.}
        
        \label{fig:2_(a)Angle_variation}
    \end{subfigure}\hspace{0.06\textwidth}%
    \begin{subfigure}{0.28\textwidth}
        \centering
        \begin{tikzpicture}[scale=0.7]
            \begin{axis}[
                xlabel={Step},
                ylabel={Error (\%)},
                xlabel style={xshift=20pt, font=\fontsize{10}{12}\selectfont\bfseries},
                ylabel style={yshift=-20pt, xshift=20pt, font=\fontsize{10}{12}\selectfont\bfseries},
                xmin=0, xmax=40,
                ymin=0, ymax=10,
                ytick={2,4,6,8,10},
                grid=major,
                grid style={line width=0.3pt, draw=gray!30},
                legend style={
                    at={(0.95,1.4)}, 
                    anchor=north, 
                    font=\fontsize{9}{11}\selectfont, 
                    draw=none, 
                    fill=white, 
                    fill opacity=0.9,
                    legend cell align=left,
                    legend image code/.code={   
                        \draw[draw={rgb,255:red,15;green,82;blue,186}, line width=1pt]
                        (0cm,0cm) -- (0.6cm,0cm);  
                    },
                },
                every axis plot/.append style={thick},
                tick label style={font=\fontsize{9}{11}\selectfont}
            ]
            \addplot[name path=max_curve, draw=none] table[x=Timestep, y={Max Error}, col sep=comma] {sec/Data/error_summary.csv};
            \addplot[name path=min_curve, draw=none] table[x=Timestep, y={Min Error}, col sep=comma] {sec/Data/error_summary.csv};
        
            \addplot[color={rgb,255:red,15;green,82;blue,186}, fill opacity=0.2] fill between[of=max_curve and min_curve];
        
            \addplot[color={rgb,255:red,15;green,82;blue,186}, thick, smooth] table[x=Timestep, y={Average Error}, col sep=comma] {sec/Data/error_summary.csv};
            \addlegendentry{Mean Error}
        
            \end{axis}
        \end{tikzpicture}
        \vspace*{-8pt}
        \put(-25, 15){(\textbf{b}) Acceleration error.}
        \label{fig:2_(b)Acceleration_error}
    \end{subfigure}\hspace{0.04\textwidth}%
    \begin{subfigure}{0.28\textwidth}
        \centering
        \begin{tikzpicture}[scale=0.7]
            \begin{axis}[
                xlabel={Step},
                ylabel={Value},
                xlabel style={xshift=20pt, font=\fontsize{10}{12}\selectfont\bfseries},
                ylabel style={yshift=-20pt, xshift=20pt, font=\fontsize{10}{12}\selectfont\bfseries},
                xmin=0, xmax=40,
                ymin=0, ymax=1.5,
                ytick={0.5,1,1.5},
                grid=major,
                grid style={line width=0.3pt, draw=gray!30},
                legend style={
                    at={(0.5,1.4)}, 
                    anchor=north, 
                    font=\fontsize{9}{11}\selectfont, 
                    draw=none, 
                    fill=white, 
                    fill opacity=0.9,
                    legend cell align=left,
                    legend image code/.code={   
                        \draw[draw={rgb,255:red,15;green,82;blue,186}, line width=1pt]
                        (0cm,0cm) -- (0.6cm,0cm);  
                    },
                },
                every axis plot/.append style={thick},
                tick label style={font=\fontsize{9}{11}\selectfont}
            ]
        
            \addplot[name path=max_curve, draw=none] table[x=Timestep, y=Max, col sep=comma] {sec/Data/latent_wg_summary.csv};
            \addplot[name path=min_curve, draw=none] table[x=Timestep, y=Min, col sep=comma] {sec/Data/latent_wg_summary.csv};
        
            \addplot[color={rgb,255:red,15;green,82;blue,186}, fill opacity=0.2] fill between[of=max_curve and min_curve];
        
            \addplot[color={rgb,255:red,15;green,82;blue,186}, thick, smooth] table[x=Timestep, y=Mean, col sep=comma] {sec/Data/latent_wg_summary.csv};
            \addlegendentry{Mean Weight}
            
            \end{axis}
        \end{tikzpicture}
        \vspace*{-8pt}
        \put(-25,15){(\textbf{c}) Convergence of  $w_g$.}
        \label{fig:2_(c)Convergence_of_wg}
    \end{subfigure}
    
    \captionsetup{justification=justified, singlelinecheck=false}
    \caption{Variation of angle, error, and weight across sampling steps on Stable Diffusion v2 using DDIM with 20 steps and 20 unique prompt-latent pairs. 
    (\textbf{a}) Angle variation per step. 
    (\textbf{b}) Error between \methodabbr and the original process, with acceleration ($r = 2$) applied over $[12,38]$, quantified by the 2-norm difference in latents. 
    (\textbf{c}) $w_g$ variation, showing initial oscillations followed by \textbf{rapid convergence}.}
    \label{fig:2_Angle_error_and_convergence}
\end{figure*}
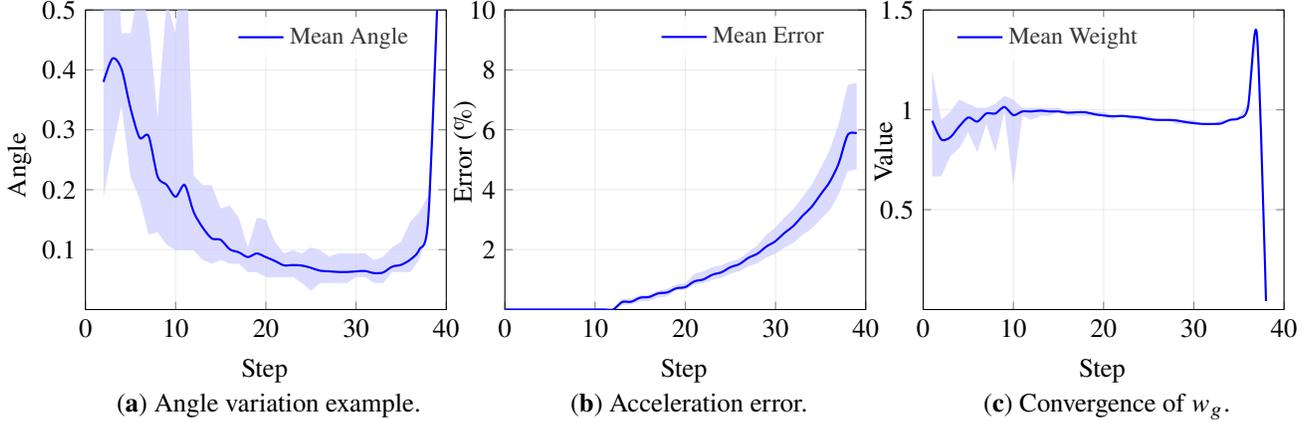

\subsection{Training-Based Acceleration Methods}

Traditional acceleration methods for diffusion models typically modify the model architecture or the sampling process during training for faster inference speed. Several typical approaches include mixed precision training and lightweight architectures~\citep{2018MobileNetV2,zhang2018shufflenet}. Recently, distillation~\citep{zhou2024score,yin2024one,sauer2024adversarial,salimans2022progressive,kim2023consistency,luo2023latent,song2023consistency} has gained popularity, focusing on reducing inference steps without significant quality loss. However, these methods require additional training, which can be both time-consuming and resource-intensive.

\subsection{Training-Free Acceleration Methods}
In certain scenarios where training resources are limited, training-free acceleration methods~\citep{karras2025guiding,qian2024boosting,xue2024accelerating,ho2022classifier} can outperform the training-based ones since they don't require any additional training, but are still capable of acceleration with little compromise to the performance. Thus, many successful training-free methods have been proposed and achieved promising effects. Optimized from the DDPM formulation, DDIM~\citep{song2020denoising} is one of the typical attempts that has reduced the number of inference steps by introducing non-Markov process. 
The DPM-Solver~\citep{zheng2023dpm,lu2022dpm,lu2022dpm+} has also achieved significant reduction in the number of inference steps, while the implementation is focused on applying the specific high-order solver to solve diffusion ODEs~\citep{zhao2023unipc,zhou2024fast,karras2022elucidating} equivalent to the denoising process. Besides, more recent approaches pay attention to preserving and reusing features from earlier steps, significantly reducing the computational load for subsequent steps. For example, DeepCache~\citep{2023DeepCache,wimbauer2024cache} has performed exceptionally by reusing cacheable high-level features from consecutive steps and only updating low-level features, thus achieving nearly lossless acceleration. Moreover, optimizations on the sampling schedules have proved to be effective as well. Align Your Steps~\citep{sabour2024align} has broadened its application scenarios by leveraging methods from stochastic calculus and applying optimal schedules specific to various solvers, pre-trained models and datasets.

\section{\methodabbr for Faster Diffusion Sampling}

In this section, we first introduce the newly observed phenomenon called \phenmn in \cref{subsec:similarity}, which reveals the statistical correlation between the outputs of adjacent transition operations in the diffusion process. Next, in \cref{subsec:single step}, we introduce our method, \methodabbr, which takes advantage of the transition operators of neighboring steps to approximate and replace the transition operator of the current step. Finally, in \cref{subsec:error_analysis}, we analyze the errors introduced by \methodabbr, which is negligible. 
\subsection{\phenmn}\label{subsec:similarity}

In this section, we introduce the newly observed phenomenon called \textbf{\phenmn}: During certain phases of the diffusion process, the transition operators of consecutive steps exhibit significant similarity. To quantify the difference between steps \( t \) and \( t+1 \), we define the transition operator as $\Delta \mathbf{x}_{t+1, t} = \mathbf{x}_t - \mathbf{x}_{t+1}.$ Additionally, we define the angle between the transition operators at successive steps as follows: 

\begin{definition}
    The angle $\theta$ between $\Delta \mathbf{x}_{t+1, t}$ and $\Delta \mathbf{x}_{t+2, t+1}$ is defined as
    \begin{equation}
        \theta = \arccos\left( \frac{\Delta \mathbf{x}_{t+1, t} \cdot \Delta \mathbf{x}_{t+2, t+1}}{\|\Delta \mathbf{x}_{t+1, t} \|^2 \|\Delta \mathbf{x}_{t+2, t+1}\|^2} \right).
    \end{equation}
\end{definition}

Note that when the angle $\theta$ approaches $0$, the update trajectories of the two transition operators are nearly identical, and can be replaced by each other. As shown in \cref{fig:2_(a)Angle_variation}, we observe that the angles are relatively small from steps $12$ to $38$, suggesting the update trajectories are highly similar during this period. This allows us to reduce unnecessary computations by approximating the current transition operator with that of the adjacent steps, thereby speeding up the sampling process significantly and effectively.

In practice, we define a threshold $\tau$ to identify the interval where $\theta$ remains relatively small, referred to as the acceleration interval and formally defined as follows:
\begin{definition}
The acceleration interval is defined as the range $[a, b]$, where $ a \geq 1$, $b \leq T$, and the value of $\theta$ for all $t \in [a, b]$ satisfies the following condition:
\begin{equation}
\label{def:Acceleration_interval}
\theta_t < \tau, \quad \forall t \in [a, b].
\end{equation}
\end{definition}
During the acceleration interval, we approximate the update trajectories of the current step using those of previous steps, allowing us to skip the calculations for these steps, as shown in \cref{fig:2_(a)Angle_variation}. Based on \phenmn, we propose a new training-free acceleration method called \methodabbr, which improves the efficiency of the sampling process while maintaining the quality of the generated samples. It’s important to note that \phenmn places no restrictions on the network’s structure, meaning that \methodabbr can be applied to nearly any diffusion model.
\subsection{\textbf{\methodabbr}: Transition Approximation}\label{subsec:single step}
\methodabbr reduces unnecessary computations by approximating the update trajectories of the current step using those from adjacent steps. In this section, we first introduce the formula for the approximated step. Next, we explain the conditions for this approximation and present the overall algorithm, followed by a description of the derivation.
\begin{definition}
    With  $\phi(t)$  denoting the denoising progress at step t, the approximated step $x_t^* $ of step $t$ is defined as
    \begin{equation}
    \label{eq:acceleration_step}
        \mathbf{x}_t^* = \mathbf{x}_{t+1} + w_g \gamma (\Delta \mathbf{x}_{t+2, t+1}),
    \end{equation}
    where $    w_g = \frac{(\Delta \mathbf{x}_{t+1, t}) \cdot (\Delta \mathbf{x}_{t+2, t+1})}{\gamma \left\| (\Delta \mathbf{x}_{t+2, t+1}) \right\|^2},\gamma = \frac{\phi(t) - \phi(t+1)}{\phi(t+1) - \phi(t+2)}.$  
\end{definition}
\textbf{Here, we would like to emphasize that although the calculation of $w_g$ relies on the target variable $\mathbf{x}_t$, $w_g$ itself is a convergent quantity that depends solely on the step $t$. }
To clearly identify the positions of the approximated steps, we define the acceleration condition:
\begin{definition}
The acceleration condition is defined as follows:
\begin{equation}
t \bmod r = r - 1,
\end{equation}
where  $t$  is the step number, and  $r$  is a constant, $\bmod$ stands for the modulo operation. 
\end{definition}




Based on these definitions, we propose \methodabbr. During inference, following \cref{def:Acceleration_interval}, we identify acceleration intervals. If the current step lies within an acceleration interval and satisfies the acceleration condition, we replace the original transition with the approximated step, as shown in \cref{eq:acceleration_step}. The detailed algorithm is presented in \cref{tab:FreeRide Acceleration}.

\subsubsection{Formalizing \texorpdfstring{$\gamma$}{gamma} and \texorpdfstring{$w_g$}{wg}}
In this section, we explain how to select \(\gamma\) and \(w_g\), which are crucial for implementing \methodabbr. The goal is to find appropriate values for \(w_g\) and \(\gamma\) that minimize the error of the approximated step, described as the difference between the approximated  value \( \mathbf{x}_t^* \) and the target value \( \mathbf{x}_t \) as follows:
\begin{definition}
    Parameter $w_g$ and $\gamma$ is defined as
    \begin{equation}
    \label{eq:optimal_values}
         w_g, \gamma = \operatorname*{argmin}_{w_g, \gamma} \left( \left\| \Delta \mathbf{x}_{t+1, t} - w_g \gamma \Delta \mathbf{x}_{t+2, t+1} \right\|^2 \right).
    \end{equation}
\end{definition}

\begin{figure}[H]
    \centering
        \begin{tikzpicture}[scale=0.9]
        \begin{axis}[
            xlabel={Bias($\times 10^{-2}$)},
            ylabel={PSNR (dB)},
            xlabel style={xshift=10pt,yshift=0pt},
            ylabel style={yshift=-10pt, xshift=15pt},
            xmin=-5, xmax=10,
            ymin=33, ymax=42,  
            xtick={-5,0,5,10},
            ytick={32,34.5,37,39.5,42},
            grid=major,
            grid style={line width=0.3pt, draw=gray!50, dotted},
            legend style={at={(0.9,1.1)}, anchor=north, font=\small, draw=none, fill=white, fill opacity=0.8},legend image post style=plot-blue,
            every axis plot/.append style={thick},
            legend image code/.code={   
                \draw[color=plot-blue, line width=1pt]
                (0cm,0cm) -- (0.6cm,0cm);  
            },
        ]
    
        \addplot[name path=max_curve, draw=none] table[x expr=\thisrow{Bias}*100, y=Max PSNR, col sep=comma] {sec/Data/psnr_summary.csv};
        \addplot[name path=min_curve, draw=none] table[x expr=\thisrow{Bias}*100, y=Min PSNR, col sep=comma] {sec/Data/psnr_summary.csv};
    
        \addplot[color=plot-blue, fill opacity=0.15, forget plot] fill between[of=max_curve and min_curve];

        \addplot[color=plot-blue, thick, smooth] table[x expr=\thisrow{Bias}*100, y=Mean PSNR, col sep=comma] {sec/Data/psnr_summary.csv};
        \addlegendentry{Mean PSNR}

        \end{axis}
        \end{tikzpicture}
        \captionsetup{justification=justified, singlelinecheck=false}
        \caption{Variation of PSNR values between original images and those generated by \methodabbr as a function of bias, following the same experimental setup as \cref{fig:2_Angle_error_and_convergence}. 
        }
        \label{fig:3_bias_psnr}
    
\end{figure}
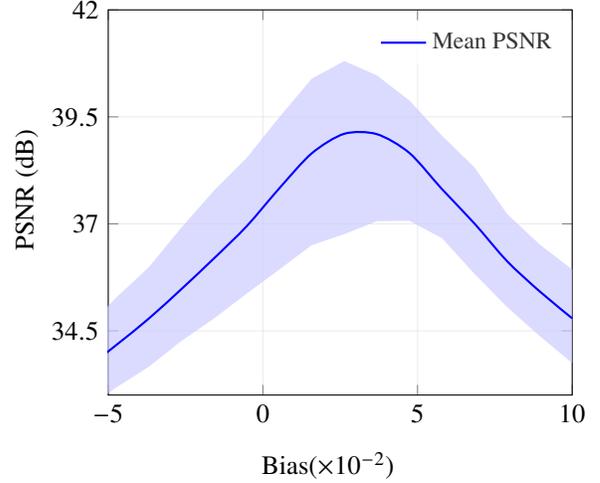

\noindent \textbf{Empirically determined value of $\gamma$:} \(\gamma\) controls the relative weighting of the step intervals, effectively adjusting the influence of denoising progress on the update process. We denote the denoising progress at step \(t\) as \(\phi(t)\) and define it as \(\phi(t) = \sqrt{\text{SNR}}_t\), where SNR represents the signal-to-noise ratio. This choice is motivated by the observation that as denoising progresses, the signal becomes increasingly dominant over noise, leading to a higher SNR. Taking the square root of SNR provides a measure that scales more linearly with the improvement in signal quality, offering a practical representation of denoising progress. Consequently, we quantify the progress over the interval \([t, t+1]\) as \(\phi(t)-\phi(t+1)\), which serves as the foundation for defining \(\gamma\) as follows: 
\begin{equation}
    \gamma = \frac{\phi(t)-\phi(t+1)}{\phi(t+1)-\phi(t+2)}.
\end{equation}

\noindent \textbf{Optimal of $w_g$:} \( w_g \) scales the magnitude of the transition operator. It is determined by minimizing the discrepancy between the actual transition and the approximated transition step, as given in \cref{eq:acceleration_step}. Here we provide the values of \( w_g \) as follows (see supplementary for details):
\begin{equation}
    w_g = \frac{\Delta \mathbf{x}_{t+1, t} \cdot \Delta \mathbf{x}_{t+2, t+1}}{\gamma \left\| \Delta \mathbf{x}_{t+2, t+1} \right\|^2}.
    \label{eq:wg}
\end{equation}

\subsubsection{Convergence analysis of \texorpdfstring{$w_g$}{wg}}
\label{subsubsec:converge}

As shown in \cref{eq:optimal_values}, the calculation of $w_g$ depends on \( \mathbf{x}_t \). However, during sampling, the value of $x_t$ is the target we aim to approximate, which poses a challenge to the approximation process. \textbf{Fortunately, we observe that $w_g$ consistently exhibits convergence for across varying \( \mathbf{x}_t \), allowing us to determine the corresponding $w_g$ solely from $t$.} In this section, we first introduce the algorithm for evaluating  $w_g$, and then employ it to conduct a rigorous convergence analysis of  $w_g$.


\noindent\textbf{Algorithm for Estimating $w_g$:}  
Computing $w_g$ is challenging because approximating $\mathbf{x}_t$ introduces cumulative errors that propagate through subsequent steps, affecting the accuracy of the sampling process, as illustrated in \cref{fig:2_(b)Acceleration_error}. To effectively mitigate this issue, we propose a two-step algorithm: 
1) compute the current $w_g$ and use it to derive the approximated step $\mathbf{x}_t^*$, which serves as the input for the next iteration; 
2) perform a local search to optimize $w_g$ across the entire acceleration interval, minimizing cumulative errors (see \cref{algorithmic:wg} for details).

\noindent\textbf{Algorithm for Refining the Estimation of $w_g$:}  
Although the $w_g$ obtained by \cref{algorithmic:wg} performs well, we introduce an optional refinement step to further improve the fidelity of accelerated images. Specifically, we use PSNR to evaluate and enhance the quality of accelerated images. This metric quantifies fidelity by measuring the similarity between an accelerated image and its non-accelerated counterpart. Typically, a PSNR above 30 dB signifies high-fidelity generation. The refinement algorithm consists of: 
1) introducing a bias to adjust all $w_g$ values, and 
2) conducting an end-to-end search within a predefined bias interval to determine the optimal adjustment (see \cref{algorithm:opwg} for details).

\noindent \textbf{Analysis of results:}  
Similar to the convergence behavior observed in~\cref{fig:2_(c)Convergence_of_wg}, the $w_g$ computed by~\cref{algorithmic:wg} consistently converges, ensuring the generality of our method (see supplementary for more details). Additionally, \cref{fig:3_bias_psnr} demonstrates that introducing a bias further enhances image similarity, increasing the PSNR from $37.5$ to $39$. 
The symmetry observed between bias and PSNR allows efficient optimization through binary search.

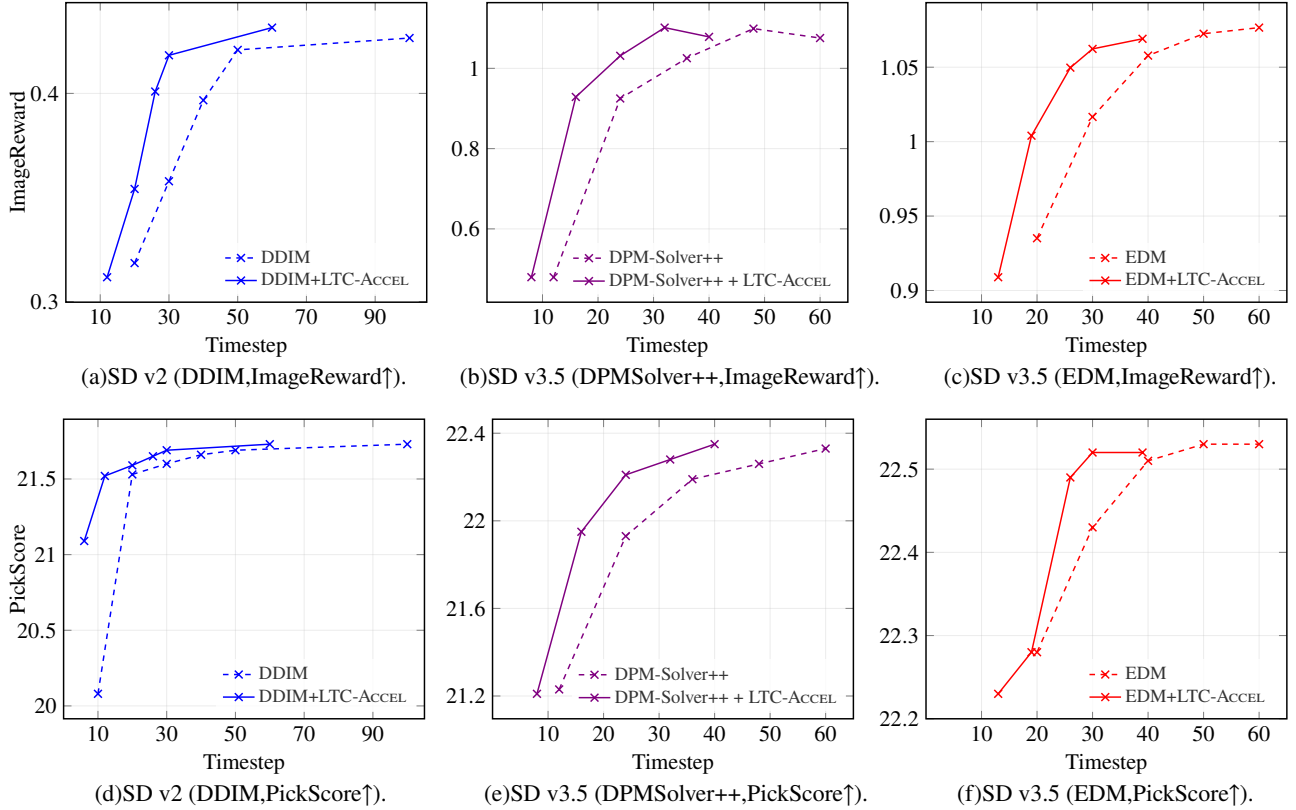
\begin{figure*}[t]
    \centering
    \begin{subfigure}{0.32\textwidth}
        \centering
        \begin{tikzpicture}[scale=0.7] 
       
            \begin{axis}[
                xlabel={Timestep},
                ylabel={ImageReward},
                ylabel style={yshift=-10pt,font=\large},
                xlabel style={font=\large},
                xmin=0,xmax=105,
                xtick={10,30,50,70,90},
                ytick={0.3,0.4},
                tick label style={font=\large},
                legend style={font=\normalsize, legend cell align={left}}, 
                legend pos=south east, 
                grid=major,
                grid style={opacity=0.3},
                legend style={at={(0.7,0.2)}, anchor=north, font=\normalsize, draw=none, fill=white, fill opacity=0.8},
                mark=x,
                line width=0.8pt,
                mark size=1pt
            ]
                \addplot[color=crystal-blue, dashed, smooth, mark=*, 
                         mark options={fill=crystal-blue, solid}] 
                table [x=x, y=y, col sep=comma] {sec/Data/sd2_ddim_org.csv}[-];
                \addlegendentry{DDIM} 

                \addplot[color=crystal-blue, smooth, mark=*, 
                         mark options={crystal-blue, solid}]  
                table [x=x, y=y, col sep=comma] {sec/Data/sd2_ddim_freeride.csv}[-];
                \addlegendentry{DDIM+\methodabbr}
            \end{axis}
        \end{tikzpicture}
        \put(-131,-8){\small(a)SD v2 (DDIM,ImageReward$\uparrow$).}
        
        \label{fig:1}
    \end{subfigure}\hspace{0.01\textwidth}%
    \begin{subfigure}{0.32\textwidth}
        \centering
        \begin{tikzpicture}[scale=0.7]
        
        \begin{axis}[
                xlabel={Timestep},
                xmin=0,xmax=65,
                xtick={10,20,30,40,50,60},
                xlabel style={font=\large},
                 tick label style={font=\large},
                grid=major,
                grid style={opacity=0.3},
                legend style={at={(0.6,0.2)}, anchor=north, font=\normalsize, draw=none, fill=white, fill opacity=0.8, legend cell align={left}},
                mark=x,
                line width=0.8pt,
                mark size=1pt
            ]
                \addplot[color=lavender-dream, dashed, mark=*, smooth,mark options={fill=lavender-dream, solid}]  
                table [x=x, y=y, col sep=comma] {sec/Data/sd35_dpm_org.csv}[-];
                \addlegendentry{DPM-Solver++} 

                \addplot[color=lavender-dream, solid, mark=*, smooth, mark options={fill=lavender-dream, solid}]  
                table [x=x, y=y, col sep=comma] {sec/Data/sd35_dpm_freeride.csv}[-];
                \addlegendentry{DPM-Solver++ + \methodabbr} 
            \end{axis}
        \end{tikzpicture}
        \put(-147,-8){\small (b)SD v3.5 (DPMSolver++,ImageReward$\uparrow$).}
        \label{fig:2}
    \end{subfigure}\hspace{0.01\textwidth}%
    \begin{subfigure}{0.32\textwidth}
        \centering
        \begin{tikzpicture}[scale=0.7]
        \begin{axis}[
                xlabel={Timestep},
                xmin=0,xmax=65,
                xtick={10,20,30,40,50,60},
                xlabel style={font=\large},
                tick label style={font=\large},
                legend style={font=\tiny, legend cell align={left}}, 
                legend pos=south east, 
                grid=major,
                grid style={opacity=0.3},
                legend style={at={(0.7,0.2)}, anchor=north, font=\normalsize, draw=none, fill=white, fill opacity=0.8},
                mark=x,
                line width=0.8pt,
                mark size=1pt
            ]
                \addplot[color=ember-glow, dashed, mark=*, smooth, mark options={fill=ember-glow, solid}]  
                table [x=x, y=y, col sep=comma] {sec/Data/sd35_euler_org.csv}[-];
                \addlegendentry{EDM}

                \addplot[color=ember-glow, solid, mark=*, smooth, mark options={fill=ember-glow, solid}]  
                table [x=x, y=y, col sep=comma] {sec/Data/sd35_euler_freeride.csv}[-];
                \addlegendentry{EDM+\methodabbr}
            \end{axis}
        \end{tikzpicture}
        \put(-130,-8){\small (c)SD v3.5 (EDM,ImageReward$\uparrow$).}
        \label{fig:3}
    \end{subfigure}

    \vspace{1em} 
    
    \begin{subfigure}{0.32\textwidth}
        \centering
        \begin{tikzpicture}[scale=0.7]
        \begin{axis}[
                xlabel={Timestep},
                ylabel={PickScore},
                ylabel style={yshift=-10pt,font=\large},
                xlabel style={font=\large},
                xmin=0,xmax=105,
                xtick={10,30,50,70,90},
                tick label style={font=\large},
                legend style={font=\small, legend cell align={left}}, 
                legend pos=south east, 
                grid=major,
                grid style={opacity=0.3},
                legend style={at={(0.7,0.2)}, anchor=north, font=\normalsize, draw=none, fill=white, fill opacity=0.8},
                mark=x,
                line width=0.8pt,
                mark size=1pt
            ]
                \addplot[color=crystal-blue, dashed, mark=*, smooth, mark options={fill=crystal-blue, solid}] 
                table [x=x, y=y, col sep=comma] {sec/Data/sd2_ddim_org_pick.csv}[-];
                \addlegendentry{DDIM} 

                \addplot[color=crystal-blue, solid, mark=*, smooth, mark options={fill=crystal-blue,solid}]  
                table [x=x, y=y, col sep=comma] {sec/Data/sd2_ddim_freeride_pick.csv}[-];
                \addlegendentry{DDIM+\methodabbr}
            \end{axis}
        \end{tikzpicture}
        \put(-125,-8){\small(d)SD v2 (DDIM,PickScore$\uparrow$).}
        
        \label{fig:4}
    \end{subfigure}\hspace{0.01\textwidth}%
    \begin{subfigure}{0.32\textwidth}
        \centering
        \begin{tikzpicture}[scale=0.7]
        \begin{axis}[
                xlabel={Timestep},
                xlabel style={font=\large},
                xmin=0,xmax=65,
                xtick={10,20,30,40,50,60},
                ytick={21.2,21.6,22,22.4},
                tick label style={font=\large},
                legend style={font=\tiny, legend cell align={left}},
                legend pos=south east, 
                grid=major,
                grid style={opacity=0.3},
                legend style={at={(0.6,0.2)}, anchor=north, font=\normalsize, draw=none, fill=white, fill opacity=0.8, legend cell align={left}},
                mark=x,
                line width=0.8pt,
                mark size=1pt
            ]
                \addplot[color=lavender-dream, dashed, smooth, mark=*, mark options={fill=lavender-dream, solid}]  
                table [x=x, y=y, col sep=comma] {sec/Data/sd35_dpm_org_pick.csv}[-];
                \addlegendentry{DPM-Solver++} 

                \addplot[color=lavender-dream, solid, mark=*, smooth, mark options={fill=lavender-dream, solid}]  
                table [x=x, y=y, col sep=comma] {sec/Data/sd35_dpm_freeride_pick.csv}[-];
                \addlegendentry{DPM-Solver++ + \methodabbr} 
            \end{axis}
        \end{tikzpicture}
        \put(-140,-8){\small (e)SD v3.5 (DPMSolver++,PickScore$\uparrow$).}
        \label{fig:5}
    \end{subfigure}\hspace{0.01\textwidth}%
    \begin{subfigure}{0.32\textwidth}
        \centering
        \begin{tikzpicture}[scale=0.7]
        \begin{axis}[
                xlabel={Timestep},
                xmin=0,xmax=65,
                xtick={10,20,30,40,50,60},
                xlabel style={font=\large},
                tick label style={font=\large},
                legend style={font=\tiny, legend cell align={left}}, 
                legend pos=south east, 
                grid=major,
                grid style={opacity=0.3},
                legend style={at={(0.7,0.2)}, anchor=north, font=\normalsize, draw=none, fill=white, fill opacity=0.8},
                mark=x,
                line width=0.8pt,
                mark size=1pt
            ]
                \addplot[color=ember-glow, dashed, mark=*, smooth, mark options={fill=ember-glow, solid}]  
                table [x=x, y=y, col sep=comma] {sec/Data/sd35_euler_org_pick.csv}[-];
                \addlegendentry{EDM}

                \addplot[color=ember-glow, solid, mark=*, smooth, mark options={fill=ember-glow, solid}]  
                table [x=x, y=y, col sep=comma] {sec/Data/sd35_euler_freeride_pick.csv}[-];
                \addlegendentry{EDM+\methodabbr}
            \end{axis}
        \end{tikzpicture}
        \put(-125,-8){\small (f)SD v3.5 (EDM,PickScore$\uparrow$).}
       
        \label{fig:6}
    \end{subfigure}
    \captionsetup{justification=justified, singlelinecheck=false}
    \caption{Quantitative results of text-to-image. We present our results on \textbf{Stable Diffusion v2 and Stable Diffusion v3.5} by measuring the ImageReward and PickScore using $1000$ prompt-image pairs. To demonstrate our compatibility with most schedulers, we use DDIM for sampling on Stable Diffusion v2, DPM-Solver++ and EDM for sampling on Stable Diffusion v3.5. The results demonstrate that our method achieves a $\mathbf{1.67\times}$ \textbf{acceleration }on Stable Diffusion v3.5, as well as a $\mathbf{1.67\times}$ \textbf{acceleration} on Stable Diffusion v2.}
    \label{fig:3_txt2img_synthesis}
\end{figure*}
\begin{algorithm}[ht]
    \caption{\methodabbr Acceleration}
    \begin{algorithmic}[1]
    
    \STATE \textbf{Input:} Diffusion model $p_{\theta}$, acceleration interval $[a, b]$, acceleration cycle $r$, mapping $\phi$, constant sequence $w_g$, initial noise $\mathbf{x}_N$.
    \STATE \textbf{Output:} $x_0$.
    \FOR{$t = N$ \textbf{to} $0$} 
        \IF{$t \in [a ,b]$ \textbf{and} $t \bmod r = r - 1$}
            \STATE $\gamma = \frac{\phi(t) - \phi(t + 1)}{\phi(t + 1) - \phi(t + 2)}$
            \STATE $\mathbf{x}_{t} = \mathbf{x}_{t + 1} + w_{g}[t] \cdot \gamma \cdot \Delta \mathbf{x}_{t+2, t+1}$
        
        \ELSE
            \STATE $\mathbf{x}_t = p_{\theta}(\mathbf{x}_{t+1},t)$
        \ENDIF
    \ENDFOR
    
    \RETURN $\mathbf{x}_{0}$
    \end{algorithmic}
    \label{tab:FreeRide Acceleration}
\end{algorithm}
\begin{algorithm}[ht]
    \caption{Calculate $w_{g}$}
    \begin{algorithmic}[1]
    \STATE \textbf{Input:} Diffusion model $p_{\theta}$, acceleration interval $[a, b]$, acceleration cycle $r$, mapping $\phi$, initial noise $\mathbf{x}_N$.
    \STATE \textbf{Output:} $w_g$.
    
    \FOR{$t = N$ \textbf{to} $0$} 
        \IF{$t \in [a ,b]$ \textbf{and} $t \bmod r = r - 1$}
            \STATE $\gamma = \frac{\phi(t) - \phi(t + 1)}{\phi(t + 1) - \phi(t + 2)}$
            
            \STATE $w_g[t] = \frac{\Delta \mathbf{x}_{t+1, t} \cdot \Delta \mathbf{x}_{t+2, t+1}}{\gamma \left\| \Delta \mathbf{x}_{t+2, t+1} \right\|^2}$
            \STATE $\mathbf{x}_t^* = \mathbf{x}_{t+1} + w_g \gamma (\Delta \mathbf{x}_{t+2, t+1})$
            \STATE $\mathbf{x}_t = \mathbf{x}_t^*$
        \ELSE
            \STATE $\mathbf{x}_t = p_{\theta}(\mathbf{x}_{t+1},t)$
        \ENDIF
    \ENDFOR
    \RETURN $w_g$
    \end{algorithmic}
    \label{algorithmic:wg}
\end{algorithm}
\begin{algorithm}[ht]
    \caption{(Optional) Improve $w_{g}$}
    \begin{algorithmic}[1]
    \STATE \textbf{Input:} Diffusion model $p_{\theta}$,free ride $p^*_{\theta}$ , bias interval $[c, d]$, noise $x_0$.
    \STATE \textbf{Output:} Constant $ bias^*$.
    \STATE $\mathbf{x}_0 = \prod_{t=N}^{0} p_{\theta}(\mathbf{x}_{t+1},t)$
    \STATE $bias^* =\underset{bias \in [c,d]}{\operatorname{argmax}}(\text{PSNR}(x_N,\prod_{t=N}^{0} p^*_{\theta}(\mathbf{x}_{t+1},t,w_g + bias) )$
    \RETURN $bias^*$
    \end{algorithmic}
    \label{algorithm:opwg}
\end{algorithm}
\subsection{Error Analysis of the Approximated Step}\label{subsec:error_analysis}

In this section, we demonstrate that the error introduced by the approximation is generally negligible. We first provide the upper bound of the error of approximating a single step, followed by a comprehensive evaluation of the actual error incurred throughout the sampling process in practice.

\noindent\textbf{Error of approximating a single step:} Using the expressions for \( w_g \) and \( \gamma \), we can estimate the relative error between the approximated step and the original step. When the angle between $\Delta \mathbf{x}_{t+1, t}$ and $\Delta \mathbf{x}_{t+2, t+1}$ is less than a threshold $\tau$, the relative error $\epsilon_r$  can be expressed as follows (see appendix for details):
\begin{equation}
    \epsilon_r = \frac{\left\|\mathbf{x}_t - \mathbf{x}^*_t\right\|^2 }{\left\|\Delta \mathbf{x}_{t+1, t}\right\|^2}
    < \tau^2.
\end{equation}

This implies that the relative error introduced by the approximate step is smaller than the update term of the original process, scaled by $\tau^2$. When $\tau$ is typically on the order of magnitude of $[0.1, 0.2]$, $\epsilon_r$ becomes negligible.


\noindent\textbf{Error in practice:} As the approximation of steps extends to \(N\) steps during inference, the error accumulates, leading to the results shown in the figure. The findings indicate that even when \(\mathbf{32.5\%}\) of the steps are approximated, the resulting error remains only \(\mathbf{6.0\%}\), demonstrating that the error is nearly imperceptible. Notably, our method, \methodabbr, achieves a PSNR of $\mathbf{36.6}$, as illustrated in \cref{fig:3_bias_psnr}.


\section{Experiments}
\label{sec:experiments}

In this section, we conduct extensive experiments to validate our method’s effectiveness. First, in \cref{sec:settings}, we provide an overview of the experimental setup. Then, in \cref{sec:image_exp} and \cref{sec:video_exp}, we present a comprehensive evaluation of text-to-image and text-to-video synthesis. In \cref{sec:other_method}, we assess the integration of our method with other methods. Finally, we conduct ablation studies and provide detailed experimental settings in the Appendix.

\subsection{Experimental Settings}
\label{sec:settings}


In all experiments, unless otherwise specified, we use a classifier-free guidance~\citep{ho2022classifier} intensity of 7.5 without additional enhancement techniques. To evaluate acceleration performance, we perform inference in float16 and measure speedup based on the number of inference steps. Additionally, we set $\phi(t)$, as mentioned in \cref{subsec:single step}, to the $\sqrt{\text{SNR}}_t$ values of the current step. All experiments are conducted without any additional training and we ensure that both the original model and \methodabbr start with consistent initial noise. 
In addition, for each prompt, we sample a unique starting noise to ensure variability. 
\begin{table*}[htbp]
\centering
\resizebox{\textwidth}{!}{%
\begin{tabular}{ccccccccc}
\toprule
\multirow{2}{*}{Model} & \multicolumn{3}{c}{DeepCache} & \multicolumn{3}{c}{\methodabbr} & \multirow{2}{*}{Acceleration Condition} & \multirow{2}{*}{Speedup} \\
\cmidrule(lr){2-4}\cmidrule(lr){5-7}
 & Steps & Time(ms) & ImageReward$\uparrow$ & Steps & Time(ms) & ImageReward$\uparrow$ & & \\
\midrule
\multirow{4}{*}{SD v2}
& 10 & 264 & -0.2246 
& 8 & 208 & -0.2739
& $t \bmod r = r - 3,\; t>4$ 
& $1.25\times$ \\
& 20 & 524 & 0.2445 
& 16 & 419 & \textbf{0.2456} 
& $t \bmod r = r - 3,\; t>8$ 
& $1.25\times$ \\
& 50 & 1411 & 0.4039 
& 38 & 1038 & \textbf{0.4096}  
& $t \bmod r = r - 3,\; t>12$ 
& $1.41\times$ \\
& 100 & 3014 & 0.4242  
& 75 & 2171 & \textbf{0.4246}
& $t \bmod r = r - 3,\; 24 < t \leq 90 $
& $1.38\times$ \\
\bottomrule
\end{tabular}%
}
\captionsetup{justification=justified, singlelinecheck=false}
\caption{Quantitative evaluation of text-to-image generation via DDIM Sampling with \textbf{\methodabbr and DeepCache}. 
In this experiment, the parameter $N$ of DeepCache is set to $N = 2$, and the period parameter $r$ is set to $r = 3$. The results are bold if they are better both in speed and quality. The results demonstrate that our method can be combined with DeepCache, achieving an \textbf{additional $\mathbf{1.41\times}$ acceleration} on top of DeepCache’s speedup.}
\label{tab:deepcache}
\end{table*}

\begin{table*}[htbp]
\centering
\setlength{\tabcolsep}{10pt} 
\adjustbox{scale=1.05}{%
\begin{tabular}{cccccc}
\toprule
\multirow{2}{*}{Model} 
& \multicolumn{2}{c}{Ays} 
& \multicolumn{2}{c}{\methodabbr} 
& \multirow{2}{*}{Speedup}\\

\cmidrule(lr){2-3} \cmidrule(lr){4-5} 
& Steps & ImageReward$\uparrow$ 
& Steps & ImageReward$\uparrow$ 
& \\
\midrule
\multirow{3}{*}{SD v1.5} 
& 10 & 0.1332
& 8 & 0.1309
& 1.25$\times$ \\
& 20 & 0.1632
& 15 & 0.1510
& 1.33$\times$ \\
& 30 & 0.1901
& 20 & 0.2131
& 1.50$\times$ \\
\bottomrule
\end{tabular}%
}
\captionsetup{justification=justified, singlelinecheck=false}
\caption{Quantitative evaluation of text-to-image generation via DPM-Solver++ sampling with \textbf{\methodabbr and Align Your Steps}. The results demonstrate that our method can be integrated with Align Your Steps, achieving a $\mathbf{1.25\times}$ \textbf{acceleration} with minimal impact on the generation quality.}
\label{tab:ays}
\end{table*}

\subsection{Text-to-image Synthesis Task}
\label{sec:image_exp}

In this section, we evaluate the performance of \methodabbr in text-to-image synthesis task.


\noindent\textbf{Settings:} We first introduce the configurations used for evaluation in this section: \textbf{1) Baselines:} DDIM for Stable Diffusion v2~\citep{Rombach_Blattmann_Lorenz_Esser_Ommer_2022}, EDM~\citep{karras2022elucidating} and DPM-Solver++~\citep{lu2022dpm} for Stable Diffusion v3.5-mid~\citep{esser2024scalingrectifiedflowtransformers}. \textbf{2) Datasets:} We employ the random $1,000$ prompts from the MS-COCO $2017$ dataset~\citep{lin2014microsoft} for Stable Diffusion v2 trials, and Stable Diffusion v3.5-mid. \textbf{3) Metrics:} In our experiments with Stable Diffusion v2 and v3.5-mid, we used the ImageReward metric~\citep{xu2023imagereward} and PickScore~\citep{NEURIPS2023_73aacd8b}. ImageReward is a model trained on human comparisons to evaluate text-to-image synthesis, considering factors like alignment with text and aesthetic quality. PickScore, based on CLIP and trained on the Pick-a-Pic dataset, predicts user preferences for generated images.
\noindent\textbf{Results:} \cref{fig:3_txt2img_synthesis} presents the acceleration results from our experiments. \methodabbr demonstrates consistently strong performance across different text-to-image models and sampling methods. In particular, we obtain a 1.67$\times$ speedup on Stable Diffusion v2 and v3.5, which is notably high for a training-free method.

\subsection{Text-to-video Synthesis Task}
\label{sec:video_exp}

In this section, we evaluate the performance of \methodabbr on several video models in the text-to-video synthesis task.

\noindent\textbf{Settings:} The configurations used for evaluation are as follows: \textbf{1) Baselines}: We use Animated-Diff~\citep{guo2023animatediff} and CogVideoX~\citep{yang2024cogvideox} $2$B as baselines, with epiCRealism~\citep{civitai25694} as the base model for Animated-Diff. \textbf{2) Datasets:} We randomly select $100$ prompts from the MS-COCO $2017$ dataset~\citep{lin2014microsoft}. \textbf{3) Metrics:} Following prior work~\citep{wu2023tune}, we evaluate video quality using Frame Consistency and Textual Faithfulness. Specifically, for Frame Consistency, we compute CLIP image embeddings for each frame and report the average cosine similarity across all frame pairs. For Textual Faithfulness, we compute the average ImageReward score between each frame and its corresponding text prompt.

\noindent\textbf{Results:} As shown in \cref{tab:text-to-video}, \methodabbr obtains promising results across different video generation models. Notably, it can even achieve a $1.54\times$ speedup with almost no impact on video generation quality.

\begin{table*}[htbp]
\centering
\adjustbox{scale=0.95}{%
\begin{tabular}{cccccccc}
\toprule
\multirow{2}{*}{Model} 
& \multicolumn{3}{c}{Original} 
& \multicolumn{3}{c}{\methodabbr} 
& \multirow{2}{*}{Speedup} \\
\cmidrule(lr){2-4} \cmidrule(lr){5-7}
& Steps & Text$\uparrow$ & Smooth$\uparrow$
& Steps & Text$\uparrow$ & Smooth$\uparrow$
& \\
\midrule
Animated-Diff 
& 10 & 0.2439 & 0.9713
& 7 & 0.2426 & 0.9700
& 1.43$\times$ \\
Animated-Diff
& 20 & 0.3050 & 0.9729
& 13 & 0.2939 & \textbf{0.9732}
& 1.54$\times$ \\
Animated-Diff
& 30 & 0.3662 & 0.9676
& 19 & 0.3465 & \textbf{0.9681}
& 1.58$\times$ \\ \midrule 
CogVideoX 2B
& 20 & -0.1441 & 0.9442
& 14 & -0.1673 & 0.9361
& 1.43$\times$ \\
CogVideoX 2B
& 30 & 0.2302 & 0.9464
& 19 & \textbf{0.2320} & 0.9435
& 1.58$\times$ \\
CogVideoX 2B
& 40 & 0.3918 & 0.9514
& 26 & 0.3775 & 0.9511
& 1.54$\times$ \\ \midrule 
CogVideoX 2B (int8)
& 40 & 0.2113 & 0.9456
& 26 & 0.2010 & 0.9449
& 1.54$\times$ \\
\bottomrule
\end{tabular}
}
\captionsetup{justification=justified, singlelinecheck=false}
\caption{Quantitative results of text-to-video, comparing \textbf{Original and \methodabbr} settings. Text and Smooth denote Textual Faithfulness and Frame Consistency, respectively. The results are bold if they are better both in speed and quality. The results demonstrate that our method can be combined with video models well, achieving a \textbf{ $\mathbf{1.58\times}$ acceleration} at most.}
\label{tab:text-to-video}
\end{table*}

\begin{table*}[htbp]
\centering
\adjustbox{scale=0.9}{%
\begin{tabular}{ccccccccccc}
\toprule
\multirow{2}{*}{Model} 
& \multicolumn{3}{c}{Original}
& \multicolumn{3}{c}{Distillation}
& \multicolumn{3}{c}{\methodabbr}
& \multirow{2}{*}{Speedup} \\
\cmidrule(lr){2-4}\cmidrule(lr){5-7}\cmidrule(lr){8-10}
& Steps & Text$\uparrow$ & Smooth$\uparrow$
& Steps & Text$\uparrow$ & Smooth$\uparrow$
& Steps & Text$\uparrow$ & Smooth$\uparrow$
& \\
\midrule
Animated-Diff-Lightning 
& 4 & 0.3662 & 0.9685
& 3 & 0.2913 & 0.9673
& 3 & 0.3550 & 0.9645
& 1.33$\times$ \\
Animated-Diff-Lightning
& 8 & 0.3371 & 0.9697
& 5 & 0.2978 & 0.9690
& 5 & \textbf{0.3493} & 0.9654
& 1.60$\times$ \\
\bottomrule
\end{tabular}
}
\captionsetup{justification=justified, singlelinecheck=false}
\caption{Quantitative results of text-to-video, comparing \textbf{Animated-Diff-Lightning (original steps), Animated-Diff-Lightning (same steps as \methodabbr)}, and \methodabbr. The results are bold if they are better both in speed and quality. The results demonstrate that our method can be integrated with Animated-Diff-Lightning, achieving an \textbf{additional $\mathbf{1.60\times}$ acceleration }with minimal impact on the generation quality.}
\label{tab:distill}
\end{table*}

\subsection{Compatibility with Other Methods}
\label{sec:other_method}
Since \methodabbr leverages the relationship between the network's output, it introduces a completely new approach to acceleration. This makes it compatible with a wide variety of both training-based and training-free methods. In this section, we combine \methodabbr with various existing acceleration methods, both training-free and training-based, for text-to-image and text-to-video tasks.

\subsubsection{Combined with Training-Free Methods}
\label{sec:1.2}

In this part, our sampler configurations are meticulously set up as follows:
\textbf{1) Baselines:} We select DeepCache, Align Your Steps, and INT$8$ quantization as representatives of training-free methods and conduct experiments on Stable Diffusion v2. Specifically, DeepCache is a novel training-free method that accelerates diffusion models by leveraging temporal redundancy in the denoising process to cache and reuse features, while Align Your Steps is a principled method for optimizing sampling schedules in diffusion models, particularly efficient in few-step synthesis. INT$8$ quantization converts high-precision model parameters to $8$-bit integers, accelerating video processing and inference while maintaining acceptable quality. \textbf{2) Datasets:} The random $1,000$ prompts from the MS-COCO $2017$. \textbf{3) Metrics:} ImageReward introduced in \cref{sec:image_exp}.
\noindent\textbf{Results:} \cref{tab:deepcache} shows combining our method with DeepCache yields a $2.3\times$ speedup and boosts ImageReward. \cref{tab:ays} shows $10$ Ays steps achieve $0.1332$, while our method reaches $0.1309$ with only $8$ steps ($1.25\times$ faster). At $20$ Ays steps ImageReward is $0.1632$, our $15$-step variant scores $0.1510$ ($1.33\times$ faster). At $30$ Ays steps ImageReward is $0.1901$, while our $20$-step variant surpasses it with $0.2131$ ($1.50\times$ speedup). Finally, INT$8$ quantization on CogVideoX remains effective with our scheduler (\cref{tab:text-to-video}).
\subsubsection{Combined with Training-Based Methods}
\label{sec:2.2}

In particular, considering the setup of the distilled model, we do not use classifier-free guidance in this section. Our sampler configurations are meticulously set up as follows:
\textbf{1) Baselines:} We use Animated-Diff-Lightning as a representative of distilled models. Animated-Diff-Lighting~\citep{lin2024animatediff} is a distilled version of the Animated-Diff model, designed to retain core functionality while optimizing performance. 
Specifically, we use the $8$-step and $4$-step weights of Animated-Diff-Lightning, where the $8$-step weights correspond to $8$-step inference and the $4$-step weights correspond to $4$-step inference. 
\textbf{2) Datasets:} The random $100$ prompts from the MS-COCO $2017$ dataset~\citep{lin2014microsoft}. 
\textbf{3) Metrics:} Frame Consistency and Textual Faithfulness introduced in \cref{sec:video_exp}. 

\noindent\textbf{Results:} 
The results in \cref{tab:distill} show that our model can effectively accelerate distilled models, and even speed up the $4$-step model to $3$ steps with minimal impact on video generation quality, when compared to the original process with the same computational cost.

\section{Limitations}

Our method has two main limitations. First, its effectiveness relies on the \phenmn phenomenon, which weakens with very few sampling steps (under three). Second, while training-free, it mainly requires tuning optimal intervals, as other hyperparameters are straightforward.

\section{Acknowledgement}
This work received strong support from all co-authors. Fan Cheng provided overall guidance and supplied the computational resources. Shangwen Zhu was deeply involved throughout the entire project. Han Zhang contributed powerful theoretical insights. Zhantao Yang offered profound advice on manuscript writing and experimental design. Qianyu Peng helped design the experimental coding pipeline. Zhao Pu and Huangji Wang also proposed valuable suggestions for the paper. Also this work was supported in part by the National Key R\&D Program of China under Grant 2022YFA1005000,in part by the NSFCunder Grant 61701304.

{
    \small
    \bibliographystyle{ieeenat_fullname}
    \bibliography{main}
}
\clearpage
\pagenumbering{Roman}
\setcounter{page}{1}

\newpage

\appendix
\section*{Overview}\label{sec:app-example}

The supplementary materials consist of three sections:
\begin{itemize}
    \item \textbf{The first section provides supplementary ablation studies mentioned in the main text} (See \cref{append_sec:ablation}).
    \item \textbf{The second section is the Mathematical Derivation of \methodabbr} (see \cref{sec:der}). The overall mathematical derivation consists of two parts:
    \begin{itemize}
        \item We present the derivation process of  $w_g$  in detail (see \cref{subsec:w_g}).
        \item We introduce the derivation process for the error upper bound inequality in detail(see \cref{subsec:ineq}).
        \item We provide additional experimental evidence demonstrating the convergence of  $w_g$  and show that this convergence holds across different schedulers (see \cref{subsec:wgcg}).
    \end{itemize}
    \item \textbf{The third section details the experimental setup for the figures} (see \cref{sec:fig}).

    \item \textbf{The final section focuses on providing detailed experimental settings} (see \cref{sec:exp_details}), including three parts:
    \begin{itemize}
        \item We provide the details of the acceleration interval and the conditions for setting the approximated steps (see \cref{subsec:ai}).
        \item We specify which experiments utilized the optional algorithm and which did not (see \cref{subsec:op}).
        \item We present intuitive experimental visual results to demonstrate the effectiveness of our \methodabbr (see \cref{sec:vr}).
    \end{itemize}

\end{itemize}

\section{Ablation Studies}
\label{append_sec:ablation}

In this section, we perform ablation studies to further assess and validate the effectiveness of our method.

\noindent\textbf{Directly Skipping Steps:} 
We compare our approach with the Skipping Steps strategy within the same acceleration framework. \cref{tab:ablation} shows that our method improves computational efficiency over the original approach while preserving generation quality better than the Skipping Steps.

\begin{table}[htbp]
\centering
\adjustbox{scale=0.8}{%
\begin{tabular}{cccccc} 
\toprule
\multirow{2}{*}{Model} 
& \multirow{2}{*}{Scheduler} 
& \multicolumn{2}{c}{Skipping Steps} 
& \multicolumn{2}{c}{\methodabbr} \\
\cmidrule(lr){3-4} \cmidrule(l){5-6}
& & Steps & ImageReward$\uparrow$ & Steps & ImageReward$\uparrow$ \\
\midrule
\multirow{3}{*}{SD v2} 
& DDIM & 7  & 0.0537 & 7  & 0.1472 \\
& DDIM & 10 & 0.2003 & 10 & 0.2442 \\
& DDIM & 13 & 0.2812 & 13 & 0.3129 \\
\cmidrule{1-6} 
\multirow{3}{*}{SD v2} 
& EDM  & 7  & 0.0158 & 7  & 0.2018 \\
& EDM  & 10 & 0.2003 & 10 & 0.3171 \\
& EDM  & 13 & 0.2582 & 13 & 0.3335 \\
\bottomrule
\end{tabular}
}
\captionsetup{justification=justified, singlelinecheck=false}
\caption{Ablation study comparing \methodabbr with the \textbf{Skipping Steps} method under the same acceleration framework. The results show that \methodabbr outperforms Skipping Steps, indicating the effectiveness of \methodabbr. The results demonstrate that our method \textbf{consistently outperforms} the Skipping Steps strategy in all scenarios.
}
\label{tab:ablation}
\end{table}

\section{Mathematical Derivation}
\label{sec:der}
\subsection{Derivation of  \texorpdfstring{$w_g$}{wg} }
\label{subsec:w_g}
To derive the $w_g$, we have:
 \begin{equation}
    \label{eq:Subsitute Step}
         w_g = \arg \min \left( \left\| \Delta \mathbf{x}_{t+1, t} - w_g \gamma \Delta \mathbf{x}_{t+2, t+1} \right\|^2 \right).
\end{equation}
We expand the objective function in terms of the inner product:

\begin{align}
\begin{split}
&\left\|\Delta \mathbf{x}_{t+1,t} - w_g \gamma\, \Delta \mathbf{x}_{t+2,t+1}\right\|^2 \\
&= \left\|\Delta \mathbf{x}_{t+1,t}\right\|^2 - 2w_g \gamma \left( \Delta \mathbf{x}_{t+1,t} \cdot \Delta \mathbf{x}_{t+2,t+1} \right) \\
&\quad + w_g^2 \gamma^2 \left\|\Delta \mathbf{x}_{t+2,t+1}\right\|^2.
\end{split}
\end{align}

Taking the derivative of this expression with respect to \( w_g \) and setting it equal to zero yields:
\begin{align}
\begin{split}
&\frac{\partial}{\partial w_g}\Bigr[ \|\Delta \mathbf{x}_{t+1,t} - w_g \gamma\, \Delta \mathbf{x}_{t+2,t+1}\|^2 \Bigr] \\ 
&= -2\gamma ( \Delta \mathbf{x}_{t+1,t} \cdot \Delta\mathbf{x}_{t+2,t+1} \Bigr) + 2w_g \gamma^2 \left\|\Delta \mathbf{x}_{t+2,t+1}\right\|^2 \\
&= 0.
\end{split}
\end{align}

Rearranging this equation, we obtain:
\begin{equation}
w_g \gamma \left\|\Delta \mathbf{x}_{t+2,t+1}\right\|^2 = \Delta \mathbf{x}_{t+1,t} \cdot \Delta \mathbf{x}_{t+2,t+1}.
\end{equation}

Thus, the final expression for \( w_g \) is given by:
\begin{equation}
w_g = \frac{\Delta \mathbf{x}_{t+1,t} \cdot \Delta \mathbf{x}_{t+2,t+1}}{\gamma \left\|\Delta \mathbf{x}_{t+2,t+1}\right\|^2}.
\end{equation}

\begin{figure*}
    \centering
    \begin{subfigure}{0.4\textwidth}
        \centering
         \begin{tikzpicture}[scale=0.7]
            \begin{axis}[
                xlabel={Step},
                ylabel={Weight},
                xmin=0, xmax=20,
                xtick={0,5,10,15,20},
                ymin=0,ymax=1.3,
                ytick={0.5,0.6,0.7,0.8,0.9,1.0,1.1,1.2,1.3},
                grid=both,
                title={},
                legend style={at={(0.5,0.5)}, anchor=north},
                mark options={fill=DeepSkyBlue} 
            ]
                
                \addplot[color=DeepSkyBlue, mark options={fill=DeepSkyBlue}] table [x=Step, y=weight_1, col sep=comma] {sec/Data/ddim20fullweights_data.csv}[-]
                    node [pos=0.75, anchor=south] {}
                    ;
                \addlegendentry{$w_g^1$}

                \addplot[color=yellow, mark options={fill=yellow}] table [x=Step, y=weight_2, col sep=comma] {sec/Data/ddim20fullweights_data.csv}[-]
                    node [pos=0.75, anchor=south] {}
                    ;
                \addlegendentry{$w_g^2$}
                
                \addplot[color=purple, mark options={fill=purple}] table [x=Step, y=weight_3, col sep=comma] {sec/Data/ddim20fullweights_data.csv}[-]
                    node [pos=0.75, anchor=south] {}
                    ;
                \addlegendentry{$w_g^3$}

                \addplot[color=brown, mark options={fill=brown}] table [x=Step, y=weight_4, col sep=comma] {sec/Data/ddim20fullweights_data.csv}[-]
                    node [pos=0.75, anchor=south] {}
                    ;
                \addlegendentry{$w_g^4$}

                \addplot[color=purple, mark options={fill=purple}] table [x=Step, y=weight_5, col sep=comma] {sec/Data/ddim20fullweights_data.csv}[-]
                    node [pos=0.75, anchor=south] {}
                    ;
                \addlegendentry{$w_g^5$}
                
            \end{axis}
        \end{tikzpicture}
        \caption{DDIM $20$ $w_g$ through all steps.}
        \label{fig:append_ddim20fullweight}
    \end{subfigure}\hspace{0.1\textwidth}
    \begin{subfigure}{0.4\textwidth}
        \centering
        \begin{tikzpicture}[scale=0.7]
                \begin{axis}[
                    xlabel={Step},
                    ylabel={Weight},
                    xmin=0, xmax=5,
                    xtick={0,1,2,3,4,5},
                    ymin=0,ymax=1,
                    ytick={0.1,0.2,0.3,0.4,0.5,0.6,0.7,0.8,0.9,1},
                    grid=both,
                    title={},
                    legend style={at={(0.2,0.5)}, anchor=north},
                    mark options={fill=DeepSkyBlue} 
                ]
                    
                    \addplot[color=DeepSkyBlue, mark options={fill=DeepSkyBlue}] table [x=Step, y=weight_1, col sep=comma] {sec/Data/ddim20weights_data.csv}[-]
                        node [pos=0.75, anchor=south] {}
                        ;
                    \addlegendentry{$w_g^1$}
    
                    \addplot[color=yellow, mark options={fill=yellow}] table [x=Step, y=weight_2, col sep=comma] {sec/Data/ddim20weights_data.csv}[-]
                        node [pos=0.75, anchor=south] {}
                        ;
                    \addlegendentry{$w_g^2$}
                    
                    \addplot[color=purple, mark options={fill=purple}] table [x=Step, y=weight_3, col sep=comma] {sec/Data/ddim20weights_data.csv}[-]
                        node [pos=0.75, anchor=south] {}
                        ;
                    \addlegendentry{$w_g^3$}
    
                    \addplot[color=brown, mark options={fill=brown}] table [x=Step, y=weight_4, col sep=comma] {sec/Data/ddim20weights_data.csv}[-]
                        node [pos=0.75, anchor=south] {}
                        ;
                    \addlegendentry{$w_g^4$}
    
                    \addplot[color=purple, mark options={fill=purple}] table [x=Step, y=weight_5, col sep=comma] {sec/Data/ddim20weights_data.csv}[-]
                        node [pos=0.75, anchor=south] {}
                        ;
                    \addlegendentry{$w_g^5$}
                    
                \end{axis}
            \end{tikzpicture}
            \caption{DDIM $20$ $w_g$ through selected steps}
            \label{fig:append_ddim20weight}
    \end{subfigure}

    \begin{subfigure}{0.4\textwidth}
        \centering
         \begin{tikzpicture}[scale=0.7]
            \begin{axis}[
                xlabel={Step},
                ylabel={Weight},
                xmin=0, xmax=20,
                xtick={0,5,10,15,20},
                ymin=0,ymax=1.3,
                ytick={0.5,0.6,0.7,0.8,0.9,1.0,1.1,1.2,1.3},
                grid=both,
                title={},
                legend style={at={(0.5,0.5)}, anchor=north},
                mark options={fill=DeepSkyBlue} 
            ]
                
                \addplot[color=DeepSkyBlue, mark options={fill=DeepSkyBlue}] table [x=Step, y=weight_1, col sep=comma] {sec/Data/dpm20fullweights_data.csv}[-]
                    node [pos=0.75, anchor=south] {}
                    ;
                \addlegendentry{$w_g^1$}

                \addplot[color=yellow, mark options={fill=yellow}] table [x=Step, y=weight_2, col sep=comma] {sec/Data/dpm20fullweights_data.csv}[-]
                    node [pos=0.75, anchor=south] {}
                    ;
                \addlegendentry{$w_g^2$}
                
                \addplot[color=purple, mark options={fill=purple}] table [x=Step, y=weight_3, col sep=comma] {sec/Data/dpm20fullweights_data.csv}[-]
                    node [pos=0.75, anchor=south] {}
                    ;
                \addlegendentry{$w_g^3$}

                \addplot[color=brown, mark options={fill=brown}] table [x=Step, y=weight_4, col sep=comma] {sec/Data/dpm20fullweights_data.csv}[-]
                    node [pos=0.75, anchor=south] {}
                    ;
                \addlegendentry{$w_g^4$}

                \addplot[color=purple, mark options={fill=purple}] table [x=Step, y=weight_5, col sep=comma] {sec/Data/dpm20fullweights_data.csv}[-]
                    node [pos=0.75, anchor=south] {}
                    ;
                \addlegendentry{$w_g^5$}
                
            \end{axis}
        \end{tikzpicture}
        \caption{DPM $20$ $w_g$ through all steps.}
        \label{fig:append_dpm20fullweight}
    \end{subfigure}\hspace{0.1\textwidth}
    \begin{subfigure}{0.4\textwidth}
        \centering
        \begin{tikzpicture}[scale=0.7]
                \begin{axis}[
                    xlabel={Step},
                    ylabel={Weight},
                    xmin=0, xmax=4,
                    xtick={0,1,2,3,4},
                    ymin=0.8,ymax=1.6,
                    ytick={0.85,0.9,0.95,1,1.5},
                    grid=both,
                    title={},
                    legend style={at={(0.2,0.7)}, anchor=north},
                    mark options={fill=DeepSkyBlue} 
                ]
                    
                    \addplot[color=DeepSkyBlue, mark options={fill=DeepSkyBlue}] table [x=Step, y=weight_1, col sep=comma] {sec/Data/dpm20weights_data.csv}[-]
                        node [pos=0.75, anchor=south] {}
                        ;
                    \addlegendentry{$w_g^1$}
    
                    \addplot[color=yellow, mark options={fill=yellow}] table [x=Step, y=weight_2, col sep=comma] {sec/Data/dpm20weights_data.csv}[-]
                        node [pos=0.75, anchor=south] {}
                        ;
                    \addlegendentry{$w_g^2$}
                    
                    \addplot[color=purple, mark options={fill=purple}] table [x=Step, y=weight_3, col sep=comma] {sec/Data/dpm20weights_data.csv}[-]
                        node [pos=0.75, anchor=south] {}
                        ;
                    \addlegendentry{$w_g^3$}
    
                    \addplot[color=brown, mark options={fill=brown}] table [x=Step, y=weight_4, col sep=comma] {sec/Data/dpm20weights_data.csv}[-]
                        node [pos=0.75, anchor=south] {}
                        ;
                    \addlegendentry{$w_g^4$}
    
                    \addplot[color=purple, mark options={fill=purple}] table [x=Step, y=weight_5, col sep=comma] {sec/Data/dpm20weights_data.csv}[-]
                        node [pos=0.75, anchor=south] {}
                        ;
                    \addlegendentry{$w_g^5$}
                    
                \end{axis}
            \end{tikzpicture}
            \caption{DPM $20$ $w_g$ through selected steps.}
            \label{fig:append_dpm20weight}
    \end{subfigure}
    \captionsetup{justification=justified, singlelinecheck=false}
    \caption{Quantitative results of the variation of $w_g$. We present our results on DDIM and DPM-Solver++ in $20$ steps, with $5$ different prompts and latents. \cref{fig:append_ddim20fullweight} and \cref{fig:append_dpm20fullweight} demonstrate the original results without acceleration, where $w_g$ achieves convergence after about $12$ steps. \cref{fig:append_ddim20weight} and \cref{fig:append_dpm20weight} show the results within the acceleration inverval $[12, 20]$, where different weights are almost the same, indicating strong feature of convergence.}
    \label{fig:append_convergence}
\end{figure*}
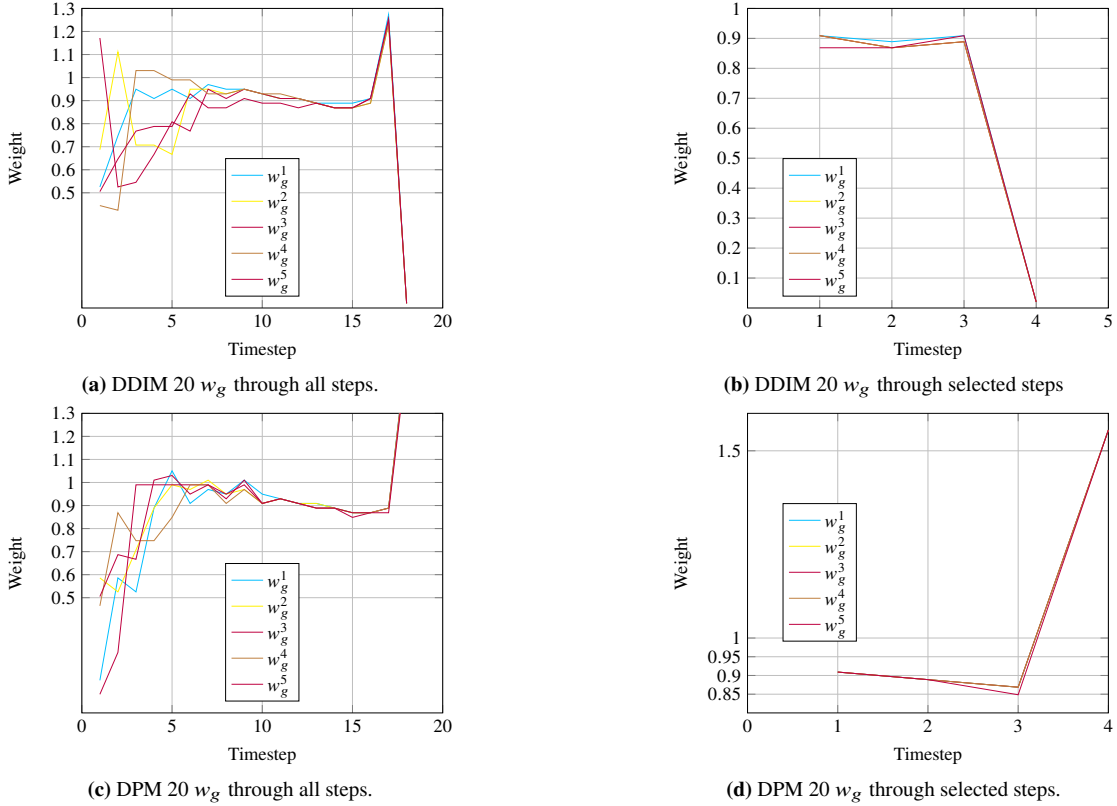

\subsection{Derivation of the Error Upper Bound Inequality}
\label{subsec:ineq}
To derive the inequality, we have:
\begin{equation}
    \theta = \arccos\left( \frac{\Delta x_{t+1, t} \cdot \Delta x_{t+2, t+1}}{\|\Delta x_{t+1, t}\|^2 \|\Delta x_{t+2, t+1}\|^2} \right) < \tau,
\end{equation}

which is equivalent to: 
\begin{equation}
    \frac{\Delta x_{t+1, t} \cdot \Delta x_{t+2, t+1}}{\|\Delta x_{t+1, t}\|^2 \|\Delta x_{t+2, t+1}\|^2} > \cos \tau.
\end{equation}
Substituting this back into \cref{eq:Subsitute Step}, we get:
\begin{equation}
    \begin{split}
        \\ \left\| \Delta x_{t+1, t}
\right\|^2 - \frac{ \left( \Delta x_{t+1, t} \cdot \Delta x_{t+2, t+1} \right)^2}{\left\| \Delta x_{t+2, t+1} \right\|^2}.
    \end{split}
\end{equation}
Squaring both sides of the inequality and substituting the expressions, we can derive:
\begin{align}
\begin{split}
&\bigl\|\Delta x_{t+1, t} - w_g\,\gamma\,\Delta x_{t+2, t+1}\bigr\|^2 \\
&\leq\bigl\|\Delta x_{t+1, t}
\bigr\|^2(1-cos^2\tau) \\
&= \bigl\|\Delta x_{t+1, t} \bigr\|^2sin^2\tau \\
&\leq\bigl\|\Delta x_{t+1, t}
\bigr\|^2\tau^2. \\
\end{split}
\end{align}
\subsection{Experimental Verification of  \texorpdfstring{$w_g$}{wg}  Convergence}
\label{subsec:wgcg}
We select DPM-Solver++~\citep{lu2022dpm}and DDIM~\citep{song2020denoising} as representative schedulers for our experiments on Stable Diffusion v2. Given that  $w_g$  tends to exhibit stronger convergence with more steps, we choose $20$ steps as a representative case.  \cref{fig:append_ddim20fullweight} and \cref{fig:append_dpm20fullweight}  illustrate the convergence behavior of the original  $w_g $, while \cref{fig:append_ddim20weight} and \cref{fig:append_dpm20weight}  demonstrate that the  $w_g$ values obtained through our algorithm also exhibit convergence.
\section{Experimental Setup for Figures}
\label{sec:fig}
In this section, we provide the experimental settings for all figures presented in the main text.
\begin{itemize}
\item \textbf{Figure 1} 
    \begin{itemize}
        \item The base model is Stable Diffusion v3.5~\citep{esser2024scalingrectifiedflowtransformers}, and the corresponding scheduler is DPM-Solver++.
        \item We select the acceleration condition as $t \bmod r = r - 1$ and $t > 4$, where the period parameter $r$ is set as $r = 2$.
    \end{itemize}
\item \textbf{Figure 2} 
    \begin{itemize}
        \item \textbf{(a)} The base model is Stable Diffusion v2~\citep{Rombach_Blattmann_Lorenz_Esser_Ommer_2022}, and the baseline is DeepCache~\citep{2023DeepCache} with $50$ steps where $N = 2$. We select the acceleration condition as $t \bmod r = r - 3$ and $t > 12$, where the period parameter $r$ is set as $r = 3$ only in experiments about DeepCache.
        \item \textbf{(b)} The baseline model is Animated-Diff~\citep{guo2023animatediff} model, and the Distillation model is Animated-Diff-Lightning~\citep{lin2024animatediff} with 4 steps under the scheduler EDM~\citep{karras2022elucidating}. We select the acceleration condition as $t \bmod r = r - 1$ and $t > 2$, where the period parameter $r$ is set as $r = 2$.
    \end{itemize}
\item \textbf{Figure 3} 
    \begin{itemize}
        \item \textbf{(a)} The base model is Stable Diffusion v2, and the corresponding scheduler is DDIM. The result is obtained at $40$ steps.
        \item \textbf{(b)} We select the acceleration condition as $t \bmod r = r - 1$ and $t > 12$ based on the original setting, where the period parameter $r$ is set as $r = 2$.
        \item \textbf{(c)} The base model is Stable Diffusion v2, and the corresponding scheduler is DDIM. The weight values are obtained at $40$ steps. Note that the acceleration condition here is  $(t > 1) $, and the approximated value is not assigned to  $x$  to prevent accumulated errors from deviating the results from the original process.
    \end{itemize}
\item \textbf{Figure 4} 
    \begin{itemize}
        \item The base model is Stable Diffusion v2, and the corresponding scheduler is DDIM. The result is obtained at $40$ steps.
        We select the acceleration condition as $t \bmod r = r - 1$ and $t > 12$ based on the original setting, where the period parameter $r$ is set as $r = 2$.
        We give trials to a series of bias from $0.0125$ to $0.05$.
    \end{itemize}
\item \textbf{Figure 5} 
    \begin{itemize}
        \item The detailed acceleration settings are available in \cref{tab:append_txt2img_sd}.
    \end{itemize}
\end{itemize}
\section{Experimental Details}
\label{sec:exp_details}

In this section, we give a further insight into relevant details of the experiments mentioned in the paper, including the settings of acceleration intervals, the optional $w_g$ algorithm.

\subsection{Acceleration Interval}
\label{subsec:ai}
As our method mentioned, we select certain consecutive timesteps as the acceleration interval if the angles formed between their transition operators are less than a threshold  $\tau$. 
Typically, we prefer to set $\tau = 0.1$. 
However, the threshold and specific acceleration interval may vary slightly by manual adjustment according to the actual angle plot, in case of a few abnormal angle data samples.

After choosing the acceleration interval, we need to set the period of acceleration, which is defined as the parameter $r$ in the paper. 
In most cases, the acceleration is applied at timestep $\mathbf{x}_t$ within the acceleration interval if $t \bmod r = r - 1$.
Generally, $r = 2$ is applicable for almost all the cases without any other modification manually, which demonstrates the versatility of our method. The acceleration condition may be adjusted if further improvements in generation quality are required.

\subsubsection*{Text-to-image Synthesis Task}
\cref{tab:append_txt2img_sd} presents the detailed results of all the experiments we conduct on Stable Diffusion v2 and v3.5-mid in the text-to-image synthesis task, with acceleration settings attached. And \cref{tab:append_txt2img_deepcache} and \cref{tab:append_txt2img_ays} show the acceleration settings of each group. 
Notably, $r = 2$ is across all these experiments, except for the case that $r = 3$ in the experiment with DeepCache in \cref{tab:append_txt2img_deepcache}. 

\begin{table*}
    \centering
    \adjustbox{scale=0.8}{%
    \begin{tabular}{ccc|cc|cc|cc}
        \toprule
          & & & \multicolumn{2}{c|}{Original} & \multicolumn{2}{c|}{\methodabbr} & & \\
        \multirow{2}{*}{Metric} & \multirow{2}{*}{Model} & \multirow{2}{*}{Scheduler} & Inference & Metric & Inference & Metric & Acceleration & \multirow{2}{*}{Speedup} \\
          & & & Step & Value & Step & Value & Condition & \\
         \midrule
        ImageReward & SD v2 & DDIM & 10 & -0.5070 & 6 & 0.0261 & $t \bmod r = r - 1$ and $t > 2$ & 1.67$\times$\\
        ImageReward & SD v2 & DDIM & 20 & 0.3185 & 12 & 0.3117 & $t \bmod r = r - 1$ and $t > 4$ & 1.67$\times$\\
        ImageReward & SD v2 & DDIM & 30 & 0.3578 & 20 & 0.3541 & $t \bmod r = r - 1$ and $t > 10$ & 1.50$\times$\\
        ImageReward & SD v2 & DDIM & 40 & 0.3967 & 26 & 0.4009 & $t \bmod r = r - 1$ and $t > 12$ & 1.54$\times$\\
        ImageReward & SD v2 & DDIM & 50 & 0.4209 & 30 & 0.4183 & $t \bmod r = r - 1$ and $t > 10$ & 1.67$\times$\\
        ImageReward & SD v2 & DDIM & 100 & 0.4266 & 60 & 0.4316 & $t \bmod r = r - 1$ and $t > 20$ & 1.67$\times$\\
        ImageReward & SD v3.5 & DPM-Solver++ & 12 & 0.4795 & 8 & 0.4796 & $t \bmod r = r - 1$ and $t > 4$ & 1.50$\times$\\
        ImageReward & SD v3.5 & DPM-Solver++ & 24 & 0.9249 & 16 & 0.9287 & $t \bmod r = r - 1$ and $t > 8$ & 1.50$\times$\\
        ImageReward & SD v3.5 & DPM-Solver++ & 36 & 1.0254 & 24 & 1.0313 & $t \bmod r = r - 1$ and $t > 12$ & 1.50$\times$\\
        ImageReward & SD v3.5 & DPM-Solver++ & 48 & 1.0990 & 32 & 1.1016 & $t \bmod r = r - 1$ and $t > 16$ & 1.50$\times$\\
        ImageReward & SD v3.5 & DPM-Solver++ & 60 & 1.0755 & 40 & 1.0785 & $t \bmod r = r - 1$ and $t > 20$ & 1.50$\times$\\
        ImageReward & SD v3.5 & EDM & 20 & 0.9351 & 13 & 0.9089 & $t \bmod r = r - 1$ and $t > 6$ & 1.53$\times$\\
        ImageReward & SD v3.5 & EDM & 30 & 1.0166 & 19 & 1.0040 & $t \bmod r = r - 1$ and $t > 8$ &1.58$\times$\\
        ImageReward & SD v3.5 & EDM & 40 & 1.0578 & 26 & 1.0497 & $t \bmod r = r - 1$ and $11 \leq t \leq 37$ & 1.54$\times$\\
        ImageReward & SD v3.5 & EDM & 50 & 1.0725 & 30 & 1.0623 & $t \bmod r = r - 1$ and $t > 10$ & 1.67$\times$\\
        ImageReward & SD v3.5 & EDM & 60 & 1.0766 & 39 & 1.0691 & $t \bmod r = r - 1$ and $15 \leq t \leq 55$ & 1.54$\times$\\
        PickScore & SD v2 & DDIM & 10 & 20.08 & 6 & 21.09 & $t \bmod r = r - 1$ and $t > 2$ & 1.67$\times$\\
        PickScore & SD v2 & DDIM & 20 & 21.53 & 12 & 21.52 & $t \bmod r = r - 1$ and $t > 4$ & 1.67$\times$\\
        PickScore & SD v2 & DDIM & 30 & 21.60 & 20 & 21.59 & $t \bmod r = r - 1$ and $t > 10$ & 1.50$\times$\\
        PickScore & SD v2 & DDIM & 40 & 21.66 & 26 & 21.65 & $t \bmod r = r - 1$ and $t > 12$ & 1.54$\times$\\
        PickScore & SD v2 & DDIM & 50 & 21.69 & 30 & 21.69 & $t \bmod r = r - 1$ and $t > 10$ & 1.67$\times$\\
        PickScore & SD v2 & DDIM & 100 & 21.73 & 60 & 21.73 & $t \bmod r = r - 1$ and $t > 20$ & 1.67$\times$\\
        PickScore & SD v3.5 & DPM-Solver++ & 12 & 21.23 & 8 & 21.21 & $t \bmod r = r - 1$ and $t > 4$ & 1.50$\times$\\
        PickScore & SD v3.5 & DPM-Solver++ & 24 & 21.93 & 16 & 21.95 & $t \bmod r = r - 1$ and $t > 8$ & 1.50$\times$\\
        PickScore & SD v3.5 & DPM-Solver++ & 36 & 22.19 & 24 & 22.21 & $t \bmod r = r - 1$ and $t > 12$ & 1.50$\times$\\
        PickScore & SD v3.5 & DPM-Solver++ & 48 & 22.26 & 32 & 22.28 & $t \bmod r = r - 1$ and $t > 16$ & 1.50$\times$\\
        PickScore & SD v3.5 & DPM-Solver++ & 60 & 22.33 & 40 & 22.35 & $t \bmod r = r - 1$ and $t > 20$ & 1.50$\times$\\
        PickScore & SD v3.5 & EDM & 20 & 22.28 & 13 & 22.23 & $t \bmod r = r - 1$ and $t > 6$ & 1.53$\times$\\
        PickScore & SD v3.5 & EDM & 30 & 22.43 & 19 & 22.28 & $t \bmod r = r - 1$ and $t > 8$ & 1.58$\times$\\
        PickScore & SD v3.5 & EDM & 40 & 22.51 & 26 & 22.49 & $t \bmod r = r - 1$ and $11 \leq t \leq 37$ & 1.54$\times$\\
        PickScore & SD v3.5 & EDM & 50 & 22.53 & 30 & 22.52 & $t \bmod r = r - 1$ and $t > 10$ & 1.67$\times$\\
        PickScore & SD v3.5 & EDM & 60 & 22.53 & 39 & 22.52 & $t \bmod r = r - 1$ and $15 \leq t \leq 55$ & 1.54$\times$\\
        \bottomrule   
    \end{tabular}
    }
    \captionsetup{justification=justified, singlelinecheck=false}
    \caption{Text-to-image Synthesis on Stable Diffusion.}
    \label{tab:append_txt2img_sd}
\end{table*}

\begin{table*}
    \centering
    \adjustbox{scale=1}{%
    \begin{tabular}{c|ccc|ccc|cc}
        \toprule
         & \multicolumn{3}{c|}{DeepCache} & \multicolumn{3}{c|}{\methodabbr} &  & \\
        \multirow{2}{*}{Model} & Inference & \multirow{2}{*}{Time} & Image & Inference & \multirow{2}{*}{Time} & Image & Acceleration & \multirow{2}{*}{Speedup} \\
         & Step & & Reward & Step & & Reward & Condition & \\
         \midrule
         \multirow{4}{*}{SD v2} & 10 & 264 & -0.2246 & 8 & 208 & -0.2739 & $t \bmod r = r - 1$ and $t > 4$ & 1.25$\times$\\
          & 20 & 524 & 0.2445 & 16 & 419 & 0.2456 & $t \bmod r = r - 1$ and $t > 8$ & 1.25$\times$\\
          & 50 & 1411 & 0.4039 & 38 & 1038 & 0.4096 & $t \bmod r = r - 3$ and $t > 12$ & 1.41$\times$\\
          & 100 & 3014 & 0.4242 & 75 & 2171 & 0.4246 & $t \bmod r = r - 3$ and $24 < t \leq 90$ & 1.38$\times$\\
        \bottomrule   
    \end{tabular}
    }
    \captionsetup{justification=justified, singlelinecheck=false}
    \caption{Quantitative results of text-to-image, combing our method with Deepcache, where the parameter $N$ mentioned in DeepCache remains $N = 2$.}
    \label{tab:append_txt2img_deepcache}
\end{table*}

\begin{table*}
    \centering
    \adjustbox{scale=0.85}{%
    \begin{tabular}{cc|cc|cc|cc|cc}
        \toprule
          & & \multicolumn{2}{c|}{Original} & \multicolumn{2}{c|}{Ays} & \multicolumn{2}{c|}{\methodabbr} & & \\
        \multirow{2}{*}{Model} & \multirow{2}{*}{Scheduler} & Inference & Image & Inference & Image & Inference & Image & Acceleration & \multirow{2}{*}{Speedup} \\
         & & Step & Reward & Step & Reward & Step & Reward & Condition & \\
         \midrule
         SD v1.5 & DPM-Solver++ & 10 & 0.1111 & 10 & 0.1332 & 8 & 0.1309 & $t \bmod r = r - 1$ and $t > 6$ & 1.25$\times$\\
        \bottomrule   
    \end{tabular}
    }
    \captionsetup{justification=justified, singlelinecheck=false}
    \caption{Quantitative results of text-to-image, combing our method with Align Your Steps.}
    \label{tab:append_txt2img_ays}
\end{table*}

\subsubsection*{Text-to-video Synthesis Task}
 We evaluate video quality from two perspectives: Frame Consistency and Textual Faithfulness. For Frame Consistency, we compute CLIP image embeddings for every frame of the output video and report the average cosine similarity among all frame pairs. For Textual Faithfulness, we compute the average ImageReward score between each output video frame and its corresponding edited prompt. From the results, we achieve a $1.54\times$ speedup with almost no impact on video generation quality.

\cref{tab:append_text-to-video} gives a specific view of the acceleration settings for the experiments we conduct on video models in the text-to-video synthesis task. In addition, \cref{tab:append_distill} presents the acceleration settings in distillation as well. 
All the experiments keep consistent in $r = 2$.
\begin{table*}
    \centering
    \adjustbox{scale=0.9}{%
    \begin{tabular}{c|ccc|ccc|cc}
        \toprule
          & \multicolumn{3}{c|}{Original} & \multicolumn{3}{c|}{\methodabbr} & & \\
         \multirow{2}{*}{Model} & Inference & Image & Frame & Inference & Image & Frame & Acceleration & \multirow{2}{*}{Speedup} \\
          & Step & Reward & Consistency & Step & Reward & Consistency & Condition &  \\
         \midrule
         epiCRealism & 10 & 0.2439 & 0.9713 & 7 & 0.2426 & 0.9700 & $t \bmod r = r - 1$ and $t > 4$ & 1.43$\times$\\
         epiCRealism & 20 & 0.3050 & 0.9729 & 13 & 0.2939 & 0.9732 & $t \bmod r = r - 1$ and $t > 6$ & 1.54$\times$\\
         epiCRealism & 30 & 0.3662 & 0.9676 & 19 & 0.3465 & 0.9681 & $t \bmod r = r - 1$ and $t > 8$ & 1.58$\times$\\
         realistic-vision & 10 & 0.1142 & 0.9636 & 7 & 0.1135 & 0.9633 & $t \bmod r = r - 1$ and $t > 4$ & 1.43$\times$\\
         realistic-vision & 20 & 0.2646 & 0.9676 & 13 & 0.2683 & 0.9672 & $t \bmod r = r - 1$ and $t > 6$ & 1.54$\times$\\
         realistic-vision & 30 & 0.4046 & 0.9655 & 19 & 0.3913 & 0.9669 & $t \bmod r = r - 1$ and $t > 8$ & 1.58$\times$\\
         CogVideoX-2B & 20 & -0.1441 & 0.9442 & 14 & -0.1673 & 0.9361 & $t \bmod r = r - 1$ and $t > 8$ & 1.43$\times$\\
         CogVideoX-2B & 30 & 0.2302 & 0.9464 & 19 & 0.2320 & 0.9435 & $t \bmod r = r - 1$ and $t > 8$ & 1.58$\times$\\
         CogVideoX-2B & 40 & 0.3918 & 0.9514 & 26 & 0.3775 & 0.9511 & $t \bmod r = r - 1$ and $t > 12$ & 1.54$\times$\\
        \bottomrule   
    \end{tabular}
    }
    \captionsetup{justification=justified, singlelinecheck=false}
    \caption{Quantitative results of text-to-video. We present our results on Animated-Diff, and CogVideoX by measuring the Textual Faithfulness and Frame Consistency using $100$ prompt-video pairs. }
    \label{tab:append_text-to-video}
\end{table*}

\begin{table*}
    \centering
    \adjustbox{scale=0.7}{%
    \begin{tabular}{c|ccc|ccc|ccc|ccc}
        \toprule
         & \multicolumn{3}{c|}{Original} & \multicolumn{3}{c|}{Original} & \multicolumn{3}{c|}{\methodabbr} & & \\
        \multirow{2}{*}{Model} & Inference & Image & Frame & Inference & Image & Frame & Inference & Image & Frame & Acceleration & \multirow{2}{*}{Speedup} \\
         & Step & Reward & Consistency & Step & Reward & Consistency & Step & Reward & Consistency & Condition & \\
         \midrule
         epiCRealism & 4 & 0.3662 & 0.9685 & 3 & 0.2913 & 0.9673 & 3 & 0.3550 & 0.9645 & $t \bmod r = r - 1$ and $t > 2$ & 1.33$\times$\\
         epiCRealism & 8 & 0.3371 & 0.9697 & 5 & 0.2978 & 0.9690 & 5 & 0.3493 & 0.9654 & $t \bmod r = r - 1$ and $t > 2$ & 1.60$\times$\\
         realistic-vision & 4 & 0.2412 & 0.9639 & 3 & 0.1249 & 0.9641 & 3 & 0.2156 & 0.9618 & $t \bmod r = r - 1$ and $t > 2$ & 1.33$\times$\\
         realistic-vision & 8 & 0.2469 & 0.9623 & 5 & -0.0095 & 0.9635 & 5 & 0.2237 & 0.9598 & $t \bmod r = r - 1$ and $t > 2$ & 1.60$\times$\\
        \bottomrule   
    \end{tabular}
    }
    \captionsetup{justification=justified, singlelinecheck=false}
    \caption{Quantitative results of text-to-video, combing our method with Animated-Diff-Lightning (the distilled version of Animated-Diff).}
    \label{tab:append_distill}
\end{table*}


\subsubsection*{Ablation Study}

The acceleration framework in the ablation experiment with "Skipping Steps" strategy is shown in \cref{tab:append_ablation}. In addition, we conduct further ablation studies on Align Your Steps and present results together with acceleration settings in \cref{tab:append_ablation_ays}.

\begin{table*}[htbp]
\centering
\adjustbox{scale=1}{%
\begin{tabular}{ccccccc} 
\toprule
\multirow{2}{*}{Model} 
& \multirow{2}{*}{Scheduler} 
& \multicolumn{2}{c}{Skipping Steps} 
& \multicolumn{2}{c}{\methodabbr} & \multirow{2}{*}{Acceleration Condition} \\
\cmidrule(lr){3-4} \cmidrule(l){5-6}
& & Steps & ImageReward$\uparrow$ & Steps & ImageReward$\uparrow$ & \\
\midrule
\multirow{3}{*}{SD v2} 
& DDIM & 7  & 0.0537 & 7  & 0.1472 & $t \bmod r = r - 1$ and $t > 4$\\
& DDIM & 10 & 0.2003 & 10 & 0.2442 & $t \bmod r = r - 1$ and $t > 6$\\
& DDIM & 13 & 0.2812 & 13 & 0.3129 & $t \bmod r = r - 1$ and $t > 6$\\
\cmidrule{1-6} 
\multirow{3}{*}{SD v2} 
& EDM  & 7  & 0.0158 & 7  & 0.2018 & $t \bmod r = r - 1$ and $t > 4$\\
& EDM  & 10 & 0.2003 & 10 & 0.3171 & $t \bmod r = r - 1$ and $t > 6$\\
& EDM  & 13 & 0.2582 & 13 & 0.3335 & $t \bmod r = r - 1$ and $t > 6$\\
\bottomrule
\end{tabular}
}
\captionsetup{justification=justified, singlelinecheck=false}
\caption{Ablation study comparing \methodabbr with the Skipping Steps method, where Skipping Steps maintains the same acceleration positions as ours. $r = 2$ is consistent in the ablation study.}
\label{tab:append_ablation}
\end{table*}

\begin{table*}[t]
    \centering
    \adjustbox{valign=c, scale=1}{%
    \begin{tabular}{cc|cc|cc|c}
        \toprule
         & & \multicolumn{2}{c|}{Ays + Skip-steps} & \multicolumn{2}{c}{Ays + \methodabbr} & \\
        \multirow{2}{*}{Model} & \multirow{2}{*}{Scheduler} & Inference & Image & Inference & Image & Acceleration \\
          & & Step & Reward & Step & Reward & Condition\\
         \midrule
        \multirow{1}{*}{SD v1.5} & DPM-Solver++ & 8 & 0.0820 & 8 & 0.1309 & $t \bmod r = r - 1$ and $t > 6$ \\
        \bottomrule   
    \end{tabular}
    }
    \captionsetup{justification=justified, singlelinecheck=false}
    \caption{Ablation study comparing our method with Align Your Steps (Ays), where Ays maintains the same acceleration positions as ours. $r = 2$ is consistent in the ablation study.}
    \label{tab:append_ablation_ays}
\end{table*}

\subsection{Optional \texorpdfstring{$w_g$}{wg} Algorithm}
\label{subsec:op}
The optional $w_g$ algorithm is designed to further minimize the difference between the $w_g$ we obtain and the optimal one, since in each iteration we just compute an approximate solution.
However, most of the experiments demonstrate that even without the optional $w_g$ algorithm our method can achieve promising effects on various models and schedulers, with little bias from the original process. Therefore, we apply the optional algorithm only to DDIM.


\subsection{Hyperparameters}
\label{subsec:hyper}

\begin{figure*}[hbt]
    \centering
    \begin{subfigure}{0.49\textwidth}
        \begin{tikzpicture}[scale=0.83]
        \begin{axis}[
            xlabel={Step},
            ylabel={Metric},
            xlabel style={xshift=3pt},
            ylabel style={yshift=-5pt, xshift=2pt},
            xmin=10, xmax=45,
            xtick={20,30,40},
            ytick = {0.5, 0.6, 0.7, 0.8, 0.9, 1.0},
            grid=major,
            grid style={gray!30, loosely dashed}, 
            legend style={at={(0.35,0.8)}, anchor=north, font=\small, draw=none, fill=white, fill opacity=0.8},
            axis line style={line width=1pt},
            tick style={line width=0.8pt},
            every axis plot/.append style={line width=1.2pt, smooth, line cap=round}
        ]
         
        \addplot[color=myorange!80!black,dash dot dot,dash phase=2pt,
        ] table[x=Step,y=Imaging_Quality,col sep=comma
        ]{Rebuttal_Fig_and_Data/video_org.csv};
        \addlegendentry{Org  (Imaging Quality)}

        \addplot[color=yellow!80!black, dash dot dot, dash phase=2pt] 
            table[x=Step, y=Temporal_Flickering, col sep=comma]{Rebuttal_Fig_and_Data/video_org.csv};
        \addlegendentry{Org (Temporal Flickering)}

        \addplot[color=myorange!80!black, line width=1.8pt, dash pattern=on 0pt off 0pt, mark=triangle*, 
        mark options={solid, fill=myorange!90!black}] 
            table[x=Step, y=Imaging_Quality, col sep=comma]{Rebuttal_Fig_and_Data/video_ltc.csv};
        \addlegendentry{Ours (Imaging Quality)}

        \addplot[color=yellow!80!black, line width=1.8pt, dash pattern=on 0pt off 0pt, mark=triangle*, 
        mark options={solid, fill=myorange!90!black}] 
            table[x=Step, y=Temporal_Flickering, col sep=comma]{Rebuttal_Fig_and_Data/video_ltc.csv};
        \addlegendentry{Ours (Temporal Flickering)}
    
        \end{axis}
        \end{tikzpicture}
        \vspace{-5pt}
        \captionsetup{justification=justified, singlelinecheck=true}
        \caption{\methodabbr on CogVideoX.}
        \label{fig:sup_video_iq}
    \end{subfigure}
    \hfill
    \begin{subfigure}{0.49\textwidth}
        \begin{tikzpicture}[scale=0.83]
        \begin{axis}[
            xlabel={Step},
            ylabel={ImageReward},
            xlabel style={xshift=3pt},
            ylabel style={yshift=-8pt, xshift=8pt},
            xmin=5, xmax=35,
            xtick={10,20,30},
            grid=major,
            grid style={gray!30, loosely dashed},
            legend style={at={(0.2,0.95)}, anchor=north, font=\small, draw=none, fill=white, fill opacity=0.8},
            axis line style={line width=1pt},
            tick style={line width=0.8pt},
            every axis plot/.append style={line width=1.2pt, smooth, line cap=round}
        ]

        \addplot[color=mygreen!80!black,dash dot dot,dash phase=2pt,] coordinates{(10,0.1332)(20,0.16318321711494355)(30,0.1901330420449376)};
        \addlegendentry{AYS}

        \addplot[color=mygreen!80!black, line width=1.8pt, dash pattern=on 0pt off 0pt, mark=triangle*, 
        mark options={solid, fill=mygreen!90!black}] coordinates{(8,0.1309)(15,0.16936568364390406)(20,0.21305280066886917)};
        \addlegendentry{AYS+Ours}


    
        \end{axis}
        \end{tikzpicture}
        \vspace{-5pt}
        \captionsetup{justification=justified, singlelinecheck=true}
        \caption{AYS at 10, 20 and 30.}
        \label{fig:sup_ays}
    \end{subfigure}
    \hfill
    \begin{subfigure}{0.49\textwidth}
        \begin{tikzpicture}[scale=0.83]
        \begin{axis}[
            ybar,
            bar width=6pt,
            xlabel={$\tau$ (Acceleration interval)},
            ylabel={ImageReward},
            xlabel style={xshift=3pt},
            ylabel style={yshift=-5pt, xshift=2pt},
            xmin=-0.15, xmax=0.45,
            xtick={0.15,0.2,0.25,0.4},
            grid=major,
            grid style={gray!30, loosely dashed}, 
            legend style={at={(0.2,0.95)}, anchor=north, font=\small, draw=none, fill=white, fill opacity=0.8},
            axis line style={line width=1pt},
            tick style={line width=0.8pt},
            every axis plot/.append style={line width=1.2pt, smooth, line cap=round}
        ]
         
        \addplot[ybar,draw=none,fill=Red0!95!black, bar shift=0pt] coordinates{(0.1,0.4131923627264332)};
        \addlegendentry{SD v2 Org}

        \addplot[ybar,draw=none,fill=Red1, bar shift=0pt] coordinates{(0.4,0.3352907943807077)};
        \addlegendentry{Interval [2,39]}

        \addplot[ybar,draw=none,fill=Red2, bar shift=0pt] coordinates{(0.25,0.3483917866908014)};
        \addlegendentry{Interval [6,39]}

        \addplot[ybar,draw=none,fill=Red3, bar shift=0pt] coordinates{(0.2,0.40142159098340197)};
        \addlegendentry{Interval [10,39]}

        \addplot[ybar,draw=none,fill=Red4, bar shift=0pt] coordinates{(0.15,0.4226344387261197)};
        \addlegendentry{Interval [14,39]}

        \addplot[color=mypurple!80!black, line width=1.8pt, dash pattern=on 0pt off 0pt, mark=triangle*, 
        mark options={solid, fill=mypurple!90!black}] coordinates{(0.4,0.3352907943807077)(0.25,0.3483917866908014)(0.2,0.40142159098340197)
        (0.15,0.4226344387261197)};
        \addlegendentry{Variation}
    
        \end{axis}
        \end{tikzpicture}
        \vspace{-5pt}
        \captionsetup{justification=justified, singlelinecheck=true}
        \caption{$\tau$ selection at 40.}
        \label{fig:sup_interval}
    \end{subfigure}
    \hfill
    \begin{subfigure}{0.49\textwidth}
        \begin{tikzpicture}[scale=0.83]
        \begin{axis}[
            ybar,
            bar width=6pt,
            ybar=0.5pt, 
            xlabel={Step},
            ylabel={ImageReward},
            xlabel style={xshift=3pt},
            ylabel style={yshift=-8pt, xshift=8pt},
            xmin=30, xmax=45,
            xtick={30,35,40},
            grid=major,
            grid style={gray!30, loosely dashed},
            legend style={at={(0.3,0.95)}, anchor=north, font=\scriptsize, draw=none, fill=white, fill opacity=0.8},
            axis line style={line width=1pt},
            tick style={line width=0.8pt},
            every axis plot/.append style={line width=1.2pt, smooth, line cap=round}
        ]




        \addplot[ybar,draw=none,fill=Blue1!50!white, bar shift=-10pt] coordinates{(40,0.3352907943807077)};
        \addlegendentry{Interval [2,39], Bias=0}



        \addplot[ybar,draw=none,fill=Blue2, bar shift=-4pt] coordinates{(40,0.41347503241128286)};
        \addlegendentry{Interval [2,39], Bias=0.05}

        \addplot[ybar,draw=none,fill=Blue4, bar shift=4pt] coordinates{(40,0.4226344387261197)};
        \addlegendentry{Interval [14,39],$\phi=\sqrt{SNR(t)}$}
        \addplot[ybar,draw=none,fill=Blue6, bar shift=10pt] coordinates{(40,0.40184151700540677)};
        \addlegendentry{Interval [14,39],$\phi'=SNR(t)$}


    
        \end{axis}
        \end{tikzpicture}
        \vspace{-5pt}
        \captionsetup{justification=justified, singlelinecheck=true}
        \caption{$\phi$ and bias at 40. }
        \label{fig:sup_phi+bias}
    \end{subfigure}
    \vspace{-10pt}
    \caption{Comparative experiments on video generations, compatibility with AYS, and ablation studies on interval selection, $\phi$ and bias.}
    \vspace{-22pt}
    \label{fig:sup_re_exp}
    
\end{figure*}

In our method, we emphasize that no manual tuning is required during deployment beyond choosing the acceleration interval—all other key hyperparameters are either automatically computed or empirically robust across different prompts and settings.

\begin{description}[itemsep=6pt]
\item[1. $w_g$]:
The main hyperparameter $w_g$ is automatically computed via Algorithm $2$ and consistently converges within the acceleration interval (Figure $3$c), indicating \textbf{prompt-agnostic} behavior. In all experiments, $w_g$ is computed once from a single prompt and reused. Its overhead is \textbf{comparable to a single forward pass}. The generality of $w_g$ is supported by strong results across diverse prompts (\cref{fig:append_convergence}).

\item[2. $\phi(t)$]:
$\phi(t)$ defines the relative importance of each timestep and serves as a smooth baseline for computing $\gamma$. Though its effect is normalized out in $w_g \cdot \gamma$, it stabilizes the solution of $w_g$ by smoothing temporal weights—a design inspired by ODE solvers. This allows $w_g$ to adapt to model dynamics while keeping $\phi$ fixed and general. Ablations in \cref{fig:sup_phi+bias} show consistent performance across different $\phi$, confirming robustness.

\item[3. $r$]:
We fix $r=2$ for all settings, except when using caching, where it adapts to reuse intermediate results.

\item[4. $\tau$]:
$\tau$ sets the acceleration interval. Larger values degrade quality via error accumulation. We suggest $\tau < 0.15$ for speed–fidelity trade-off.

\item[5. Bias]: 
The bias promotes global refinement over local updates, as discussed in Section $3.2.2$, with supporting results in Figure $4$ further demonstrating this design yields clear quality gains over greedy baselines.
\end{description}

To further validate the generalizability of these hyperparameters, we evaluate $\tau$, bias, and $\phi$ as shown in \Cref{fig:sup_interval,fig:sup_phi+bias}. Results present the followings: (1) Large $\tau$ degrades quality due to unstable denoising; (2) Bias enhances global performance; (3) $\phi$'s choices minimally affect outcomes, indicating robustness.

\subsection{Additional Video Experiments}
\label{subsec:append_video_metrics}

Beyond the text-to-video experiments presented in the paper, we have conducted additional experiments using novel datasets and evaluation metrics. Specifically, we integrate vBench, a perceptual benchmark into our pipeline. To better reflect real-world scenarios, we use WebVid-style prompts, suited for video generation. We assess video quality via imaging quality and temporal flickering, as shown in \Cref{fig:sup_video_iq}. Despite substantial acceleration, flickering increases only marginally ($\approx$ 0.02), confirming that perceptual temporal consistency is largely preserved.

\section{Visual Results from Selected Experiments}
\label{sec:vr}
\begin{figure*}[th]
\vskip 0.2in
\begin{center}
\begin{overpic}[width=1.9\columnwidth]{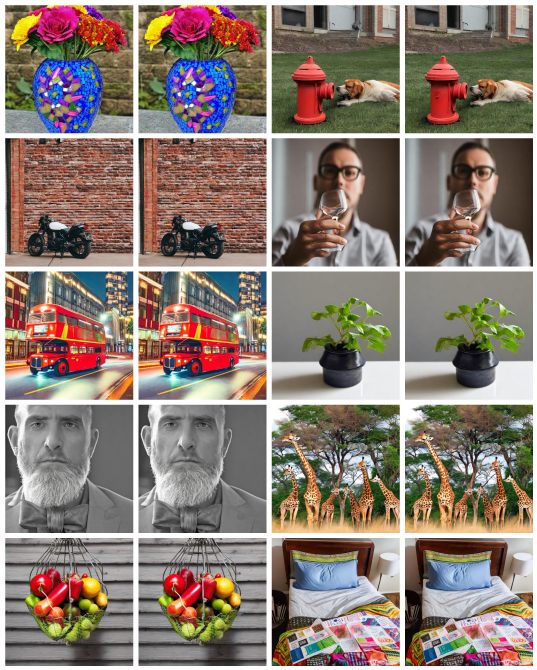}
\end{overpic}

\caption{Results obtained using DDIM sampling on Stable Diffusion v2 are shown. In each row, the first and third images correspond to the $50$-step outputs from the original process, while the remaining two images display the \methodabbr results achieved in just $30$ steps.}
\label{fig:append_DDIM50}
\end{center}
\vskip -0.2in
\end{figure*}

\begin{figure*}[th]
\vskip 0.2in
\begin{center}
\begin{overpic}[width=1.9\columnwidth]{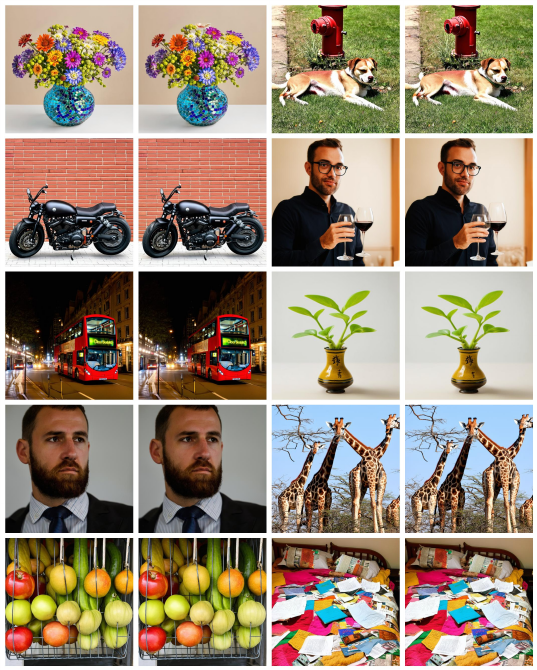}
\end{overpic}
\caption{Results obtained using EDM sampling on Stable Diffusion v3.5 are shown. In each row, the first and third images correspond to the $60$-step outputs from the original process, while the remaining two images display the \methodabbr results achieved in just $39$ steps.}
\label{fig:append_sd35}
\end{center}
\vskip -0.2in
\end{figure*}

\begin{figure*}[th]
\vskip 0.2in
\begin{center}
\begin{overpic}[width=1.9\columnwidth]{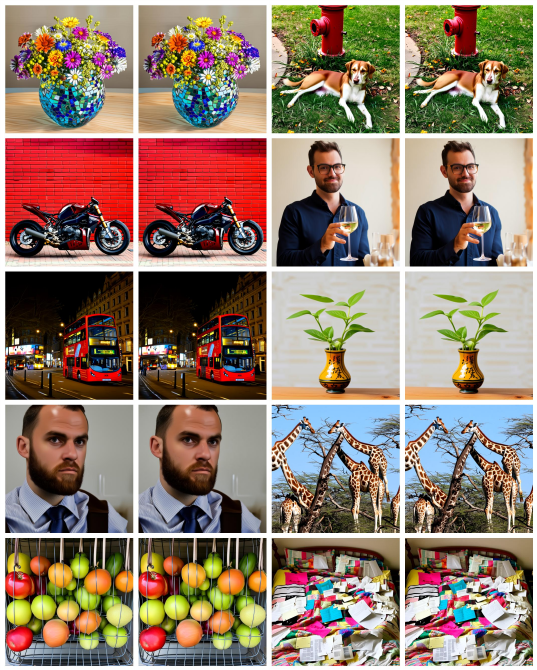}
\end{overpic}
\caption{Results obtained using DPMsolver++ sampling on Stable Diffusion v3.5 are shown. In each row, the first and third images correspond to the $60$-step outputs from the original process, while the remaining two images display the \methodabbr results achieved in $40$ steps.}
\label{fig:append_DPM50}
\end{center}
\vskip -0.2in

\end{figure*}

\begin{figure*}[th]
\vskip 0.2in
\begin{center}
\begin{overpic}[width=1.9\columnwidth]{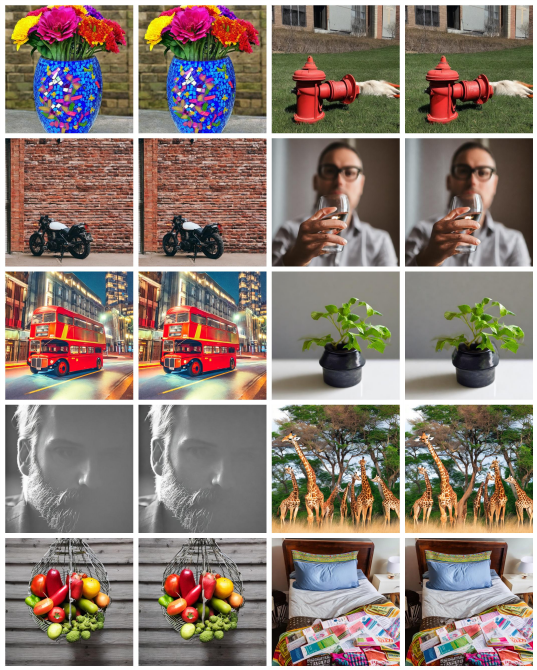}
\end{overpic}
\caption{Based on the DeepCache model, the results obtained using DDIM sampling on Stable Diffusion v2 are presented. In each row, the first and third images depict the DeepCache results after $50$ iterations, while the remaining two images display the outputs from \methodabbr combined with DeepCache after $38$ iterations.}
\label{fig:append_cache50}
\end{center}
\vskip -0.2in
\end{figure*}

\begin{figure*}[th]
\vskip 0.2in
\begin{center}
\begin{overpic}[width=1.9\columnwidth]{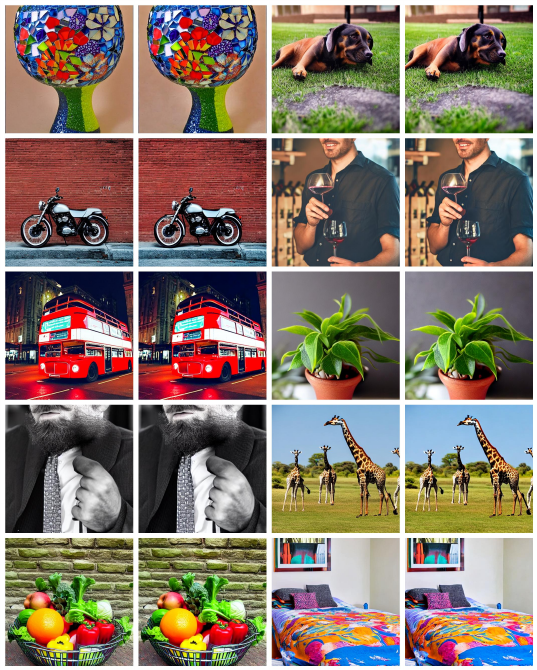}
\end{overpic}
\caption{Based on the Align Your Steps method, we obtained sampling results using DPM-Solver++ on Stable Diffusion v1.5. In each row, the first and third images represent the outputs after $10$ iterations using only the “Align Your Steps” approach, while the second and fourth images show the results achieved by combining \methodabbr with align your step for $8$ iterations.}
\label{fig:append_ays50}
\end{center}
\vskip -0.2in
\end{figure*}

\begin{figure*}[th]
\vskip 0.2in
\begin{center}
\begin{overpic}[width=1.9\columnwidth]{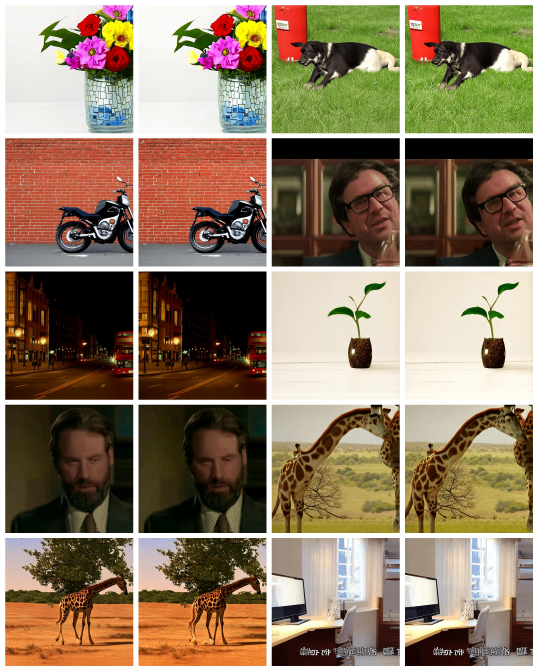}
\end{overpic}
\caption{Results obtained using DDIM sampling on CogVideoX-$2$B, with only the first frame of each video selected. In each row, the first and third images represent the outputs after $40$ iterations of the original process, while the remaining two images display the \methodabbr results achieved in $26$ iterations.}
\label{fig:append_cog}
\end{center}
\vskip -0.2in
\end{figure*}

\begin{figure*}[th]
\vskip 0.2in
\begin{center}
\begin{overpic}[width=1.9\columnwidth]{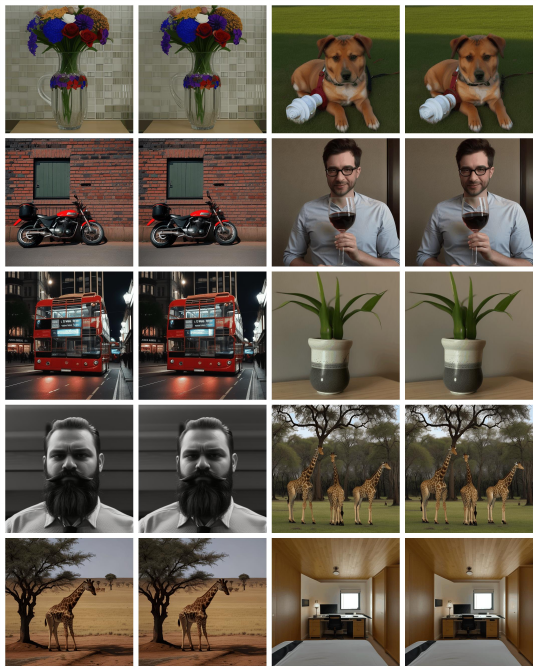}
\end{overpic}
\caption{Using EDM sampling on the Animated-Diff model based on epiCRealism, we obtained results where only the first frame of each video was selected. In each row, the first and third images correspond to the outputs after $30$ iterations of the original process, while the remaining two images show the \methodabbr results achieved in $19$ iterations.}
\label{fig:append_aep}
\end{center}
\vskip -0.2in
\end{figure*}

\begin{figure*}[th]
\vskip 0.2in
\begin{center}
\begin{overpic}[width=1.9\columnwidth]{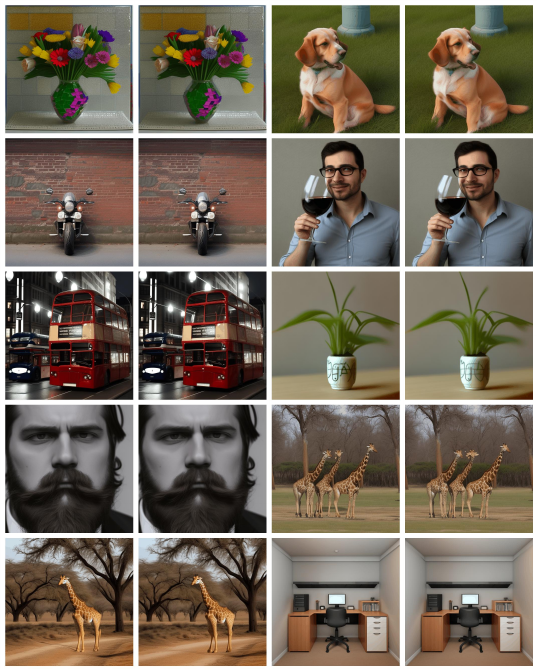}
\end{overpic}
\caption{Using EDM sampling on the Animated-Diff model based on realistic-vision, we obtained results where only the first frame of each video was selected. In each row, the first and third images correspond to the outputs after $30$ iterations of the original process, while the remaining two images show the \methodabbr results achieved in $19$ iterations.}
\label{fig:append_arp}
\end{center}
\vskip -0.2in
\end{figure*}

\begin{figure*}[th]
\vskip 0.2in
\begin{center}
\begin{overpic}[width=1.9\columnwidth]{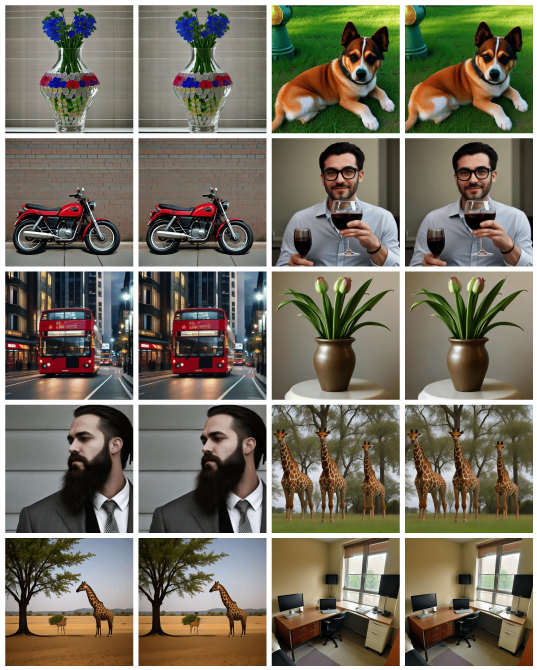}
\end{overpic}
\caption{Using EDM sampling on the Animated-Diff-Lightning model based on epiCRealism, we obtained results where only the first frame of each video was selected. In each row, the first and third images correspond to the outputs after $4$ iterations of the original process, while the remaining two images show the \methodabbr results achieved in $3$ iterations.}
\label{fig:append_alep}
\end{center}
\vskip -0.2in
\end{figure*}

\begin{figure*}[th]
\vskip 0.2in
\begin{center}
\begin{overpic}[width=1.9\columnwidth]{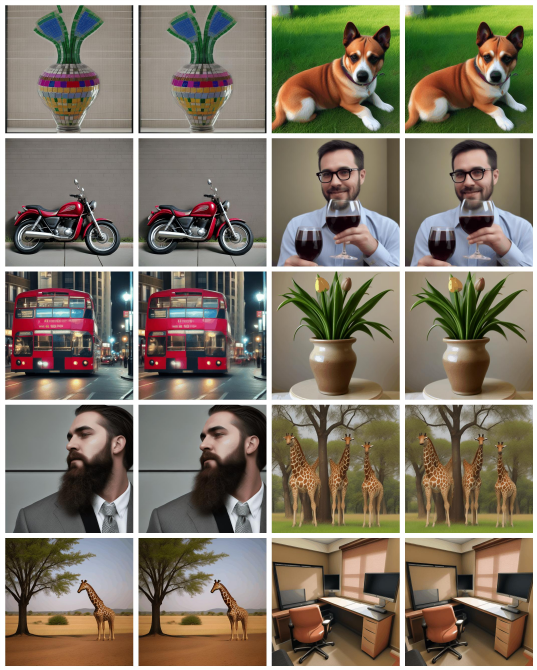}
\end{overpic}
\caption{Using EDM sampling on the Animated-Diff-Lightning model based on realistic-vision, we obtained results where only the first frame of each video was selected. In each row, the first and third images correspond to the outputs after $4$ iterations of the original process, while the remaining two images show the \methodabbr results achieved in $3$ iterations.}
\label{fig:append_alrp}
\end{center}
\vskip -0.2in
\end{figure*}

To provide a more intuitive presentation of our experimental results, we have selected representative images from our experiments for visualization. For video experiments, only the first frame is extracted for comparison. The experimental settings correspond to the displayed images as follows.
\begin{itemize}
\item \textbf{\cref{fig:append_DDIM50}} Results obtained using DDIM sampling on Stable Diffusion v2. 
\item \textbf{\cref{fig:append_sd35}} Results obtained using EDM sampling on Stable Diffusion v3.5.
\item \textbf{\cref{fig:append_DPM50}} Results obtained using DPMsolver++ sampling on Stable Diffusion v3.5.
\item \textbf{\cref{fig:append_cache50}} Based on the DeepCache model, the results obtained using DDIM sampling on Stable Diffusion v2.
\item \textbf{\cref{fig:append_ays50}} Based on the Align Your Steps method, we obtained sampling results using DPM-Solver++ on Stable Diffusion v1.5.
\item \textbf{\cref{fig:append_cog}} Results obtained using DDIM sampling on CogVideoX-2B.
\item \textbf{\cref{fig:append_aep}} Using EDM sampling on the Animated-Diff model based on epiCRealism.
\item \textbf{\cref{fig:append_arp}} Using EDM sampling on the Animated-Diff model based on realistic-vision.
\item \textbf{\cref{fig:append_alep}} Using EDM sampling on the Animated-Diff-Lightning model based on epiCRealism.
\item \textbf{\cref{fig:append_alrp}} Using EDM sampling on the Animated-Diff-Lightning model based on realistic-vision.
\end{itemize}

\end{document}